\PassOptionsToPackage{numbers}{natbib}
\documentclass{article}

\usepackage[main, final]{neurips_2026}
\usepackage[utf8]{inputenc} % allow utf-8 input
\usepackage[T1]{fontenc}    % use 8-bit T1 fonts
\usepackage{hyperref}       % hyperlinks
\hypersetup{
    colorlinks=true,
    citecolor=blue,
    linkcolor=red,
    urlcolor=blue
}
\usepackage{url}            % simple URL typesetting
\usepackage{booktabs}       % professional-quality tables
\usepackage{amsfonts}       % blackboard math symbols
\usepackage{nicefrac}       % compact symbols for 1/2, etc.
\usepackage{microtype}      % microtypography
\usepackage[table]{xcolor}
\usepackage{caption}
\usepackage{wrapfig}

\usepackage{adjustbox}
\usepackage{multirow}
\usepackage{multicol}
\usepackage{amsmath}
\usepackage{array}
\usepackage{enumitem}
\usepackage{graphicx}
\usepackage{booktabs}
\usepackage{xspace}
\usepackage[capitalize]{cleveref}
\usepackage{subcaption}

\usepackage{bbm}

\usepackage{xcolor}
\usepackage{tikz}
\usepackage{fontawesome5}
\definecolor{LAOLinkColor}{HTML}{245A81}
\definecolor{LAOCodeColor}{HTML}{30343B}
\definecolor{LAOProjectColor}{HTML}{2563A6}
\newcommand{\LAOCodeURL}{https://github.com/yaqicai/LAO-X}
\newcommand{\LAOProjectURL}{https://yaqicai.github.io/lao-x/}

\newcommand{\LAOResources}{%
  \LAOLinkButton{LAOCodeColor}{\LAOCodeURL}
    {\faGithub\hspace{0.5em}Code}%
  \hspace{0.8em}%
  \LAOLinkButton{LAOProjectColor}{\LAOProjectURL}
    {\faGlobe\hspace{0.5em}Project Page}%
}

\renewcommand{\LAOResources}{%
  \begingroup
  \hypersetup{urlcolor=LAOLinkColor}%
  \normalfont\normalsize
  \href{\LAOCodeURL}{\faGithub\hspace{0.4em}Code}%
  \hspace{1.2em}%
  \href{\LAOProjectURL}{\faGlobe\hspace{0.4em}Project Page}%
  \endgroup
}

\newcommand{\ie}{\textit{i}.\textit{e}.}
\newcommand{\eg}{\textit{e}.\textit{g}.}

\newcommand{\method}{LAO-X\xspace}

\definecolor{darkorange}{rgb}{1.0, 0.55, 0.0}
\newcommand{\cai}[1]{{#1}}

\definecolor{darkred}{rgb}{0.55, 0, 0}
\newcommand{\revise}[1]{{#1}}

\title{Localize Any Object in X-Ray Security Scans without Human Annotation}

\author{Yaqi Cai$^1$, Mingxuan Liu$^1$, Lorenzo Vaquero$^2$, Ning Wang$^3$, Nan Pu$^4$, \\
\textbf{Feng Xue$^{1}$\thanks{corresponding author}~~, Elisa Ricci$^{1,2}$, Nicu Sebe$^1$} \\
  $^1$University of Trento
  $^3$Dalian University of Technology \\
  $^2$Fondazione Bruno Kessler
  $^4$Hefei University of Technology
}

\begin{document}

\maketitle

\vspace{-20pt}
{\centering \LAOResources\par}

\begin{figure*}[h]
\centering
\includegraphics[width=1\linewidth]{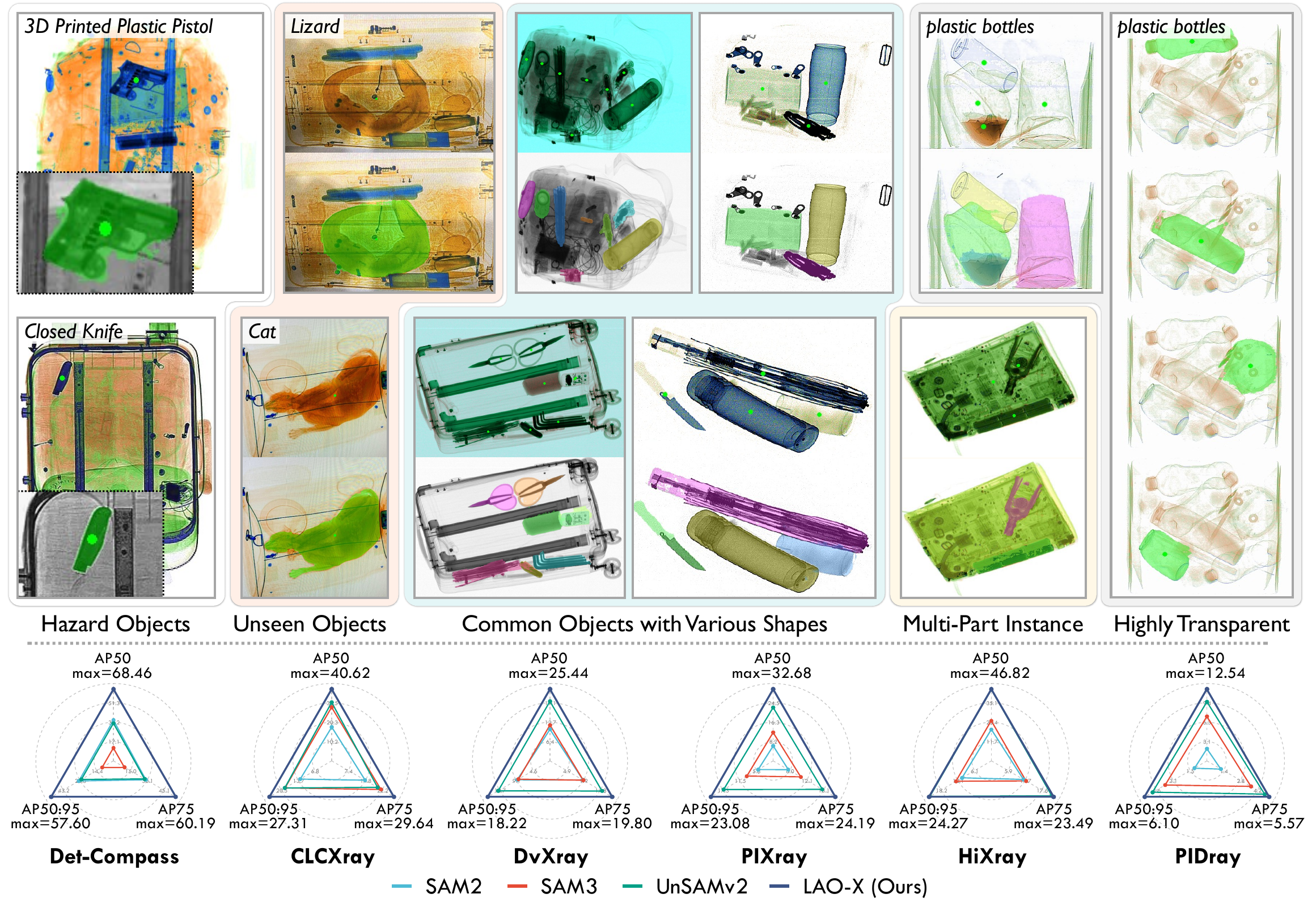}
\caption{
\textbf{Any-object localization in X-ray screening.}
\textbf{Top:} Point-prompted segmentation on challenging X-ray cases, including \emph{hazard objects}, \emph{unseen objects}, \emph{common objects with diverse shapes}, \emph{multi-part instances}, and \emph{highly transparent} items. The green dot indicates the point prompt. Different colors denote predicted masks.
\textbf{Bottom:} Radar plots of \textbf{AP50}, \textbf{AP75}, and \textbf{AP$_{50}^{95}$} on six benchmarks (Det-Compass, CLCXray, DvXray, PIXray, HiXray, PIDray), comparing SAM2, SAM3, UnSAMv2, and \textbf{LAO-X (Ours)}. LAO-X consistently achieves the strongest performance.
}
   \label{fig:teaser} % 
\end{figure*}

\begin{abstract}
Universal object localization in X-ray security inspection is critical for automated threat detection in safety-critical venues.
However, unlike everyday RGB images that dominate web-scale visual data, X-ray scans exhibit distinct color patterns, ambiguous boundaries, and compositional structures caused by volumetric superposition. 
These gaps hinder the direct zero-shot transfer of dense perception foundation models trained on web-scale RGB data. 
Moreover, annotated X-ray data is scarce and requires expert labeling, limiting both the training of generalizable X-ray native models and the adaptation of RGB foundation models for X-ray data via fine-tuning.
Given these challenges, the bright promise of highly generalizable perception models, enabled by data scaling laws in the RGB domain, remains largely out of reach for X-ray inspection.
To this end, we introduce LAO-X, a self-supervised adaptation framework that Locates Any Object in X-ray scans using diverse synthesized image--annotation pairs with granularity-aware supervision.
LAO-X first designs a saliency-guided X-ray object mining module to separate diverse object instances, which are then used for physics-guided synthesis in the absorbance domain.
LAO-X further incorporates an occlusion-controlled curriculum strategy to fine-tune a Segment Anything Model 2 (SAM2) localizer, progressively adapting it to X-ray scans with increasing object counts and overlap levels.
Experiments on six X-ray benchmarks show that LAO-X substantially improves category-agnostic localization, achieving 2\% to 23\% mAP gains over SAM2 and X-ray specific baselines in heavily cluttered scenarios, entirely without human-annotated labels.
\end{abstract}

\section{Introduction}
X-ray security screening is a critical safeguard for detecting concealed dangerous items in safety-sensitive scenarios, such as airports and large public events~\cite{Singh24}.
Although X-ray imaging enables non-intrusive inspection, it still relies on well-trained professionals to manually examine cluttered and overlapping contents. 
The consistent expansion of global mobility and logistics poses the demand for fast and reliable inspection systems, challenging operational \textit{efficiency}.
Meanwhile, advances in manufacturing technologies facilitate the creation of irregular and previously unseen threats (\eg, 3D-printed weapons), further complicating \revise{\textit{generalized}} object detection~\cite{3d_print}.

Traditional automated X-ray screening methods typically rely on fully-supervised detectors
\cite{xray_detction1_physical_scale,xray_detection2_antioverlap,dvxray}
trained on pre-defined categories.
However, the high cost of expert annotation limits the diversity of object categories and scanning
scenes, making these models brittle to unknown objects and domain shifts across scanners.
In contrast, RGB vision has entered the foundation-model era, with methods such as the Segment
Anything Model (SAM) series~\cite{sam3,sam1,sam2} pre-trained at scale to learn transferable
objectness priors and segment ``\emph{anything}'' in a zero-shot manner.
However, such RGB-based foundation models remain unreliable on X-ray imagery, where surface
textures and visible boundaries are replaced by density cues, transparency, and projective
superposition (\cref{fig:teaser}).

The generalized localization capability of SAM-based foundation models is derived from diverse
objects and comprehensive mask annotations in large-scale RGB datasets.
In X-ray security screening, however, acquiring scans requires specialized equipment, and
exhaustively annotating existing scans is impractical due to clutter, weak boundaries, severe overlap,
and scanner-dependent appearance.
As a result, existing X-ray datasets lack the category diversity and dense mask supervision required
to train a generalized localizer from human annotations.
At the same time, directly reusing RGB objectness priors is insufficient, since X-ray objects are not
defined by visible surfaces alone but by projected density patterns under physical superposition.
Facing these challenges, a central question arises: \textit{Can we adapt the ``localize any object''
capability of RGB foundation models to X-ray security screening without manual X-ray training annotations?}

We answer this question with \textbf{\method}, a self-supervised framework for \underline{L}ocating
\underline{A}ny \underline{O}bject in \underline{X}-ray scans.
\method adapts the base localizer SAM2~\cite{sam2} to the X-ray modality using physics-guided pseudo samples generated and annotated entirely from unlabeled X-ray data.
The adaptation requires no manually annotated X-ray masks, boxes, points, or category labels, while leveraging CLIP, CutLER, and SAM2 as generic pretrained priors.
Specifically, \method first mines salient foreground regions from unlabeled \revise{classification} scans and extracts object-centric masks to construct an X-ray object bank.
Then, we design physics-guided X-ray inspection synthesis to recompose the mined objects into controllable scenes in accordance with the Beer--Lambert law~\cite{swinehart1962beer}.
Since a synthesized X-ray object can provide both a reliable local component and a holistic instance extent, we organize the generated pseudo masks along a local-to-holistic granularity scale: holistic instance masks supervise object-level localization, while over-segmented local component masks provide complementary fine-structure supervision under weak boundaries and overlap.
Finally, \method trains a granularity-conditioned SAM2~\cite{unsamv2} under an occlusion-controlled curriculum~\cite{curriculum_1_weakSam}, progressively increasing scene complexity from sparse objects to severe superposition.
This design converts unlabeled X-ray scans into structured, physics-aware supervision, enabling stable category-agnostic localization on real scans without manual labels.
Extensive experiments on X-ray security screening datasets demonstrate the superiority of \method. 
Our method achieves state-of-the-art performance in both mAP and AR, surpassing baselines such as SAM3~\cite{sam3} and domain-specific open-vocabulary detectors like RAXO~\cite{raxo} by up to \textbf{23.48\%} points across objects from over 300 categories.
Qualitatively, \method produces cleaner boundaries under complex attenuation and recovers finer structures across multi-scale threats.
Overall, our contributions are \textit{threefold}:
\begin{itemize}
\item We address an underexplored foundation for X-ray security perception: localizing objects before recognizing them. Unlike closed-set or open-vocabulary detectors that rely on predefined or prompted semantics, our setting targets category-agnostic X-ray objectness without category names, text prompts, visual exemplars, or target-domain annotations.

\item We propose a systematic X-ray synthesis-and-labeling pipeline that converts unlabeled scans into physics-consistent image-annotation pairs. It mines object-centric regions, recomposes them in the Beer-Lambert absorbance domain, and generates structured local-to-holistic pseudo masks without human labels.

\item We evaluate \method on six test-only X-ray benchmarks to assess cross-dataset generalization from unlabeled source scans.
without manual X-ray training annotations, \method achieves stronger category-agnostic localization than promptable foundation models and self-supervised segmentation baselines,
and also transfers effectively to threat-item benchmarks.
\end{itemize}

\section{Related Work}

\noindent\textbf{Object Detection in X-Ray Security Screening.}
Conventional automated X-ray security inspection methods typically follow a fully supervised paradigm~\cite{xray_detction1_physical_scale,xray_detection2_antioverlap,dvxray}.
These methods operate under a closed-set assumption~\cite{sima2024multi,eds,10082660}, which learn to localize predefined categories from expert annotations.
However, existing benchmarks such as CLCXray~\cite{clcxray} and PIXray~\cite{pixray} contain fewer than 20 categories, largely due to the high cost of annotating complex X-ray scans with severe overlap.
As a result, fully supervised methods often exhibit limited generalization when encountering out-of-distribution objects and scenes.
To address evolving object distributions in luggage screening, recent works explore open-vocabulary detection with large-scale vision-language models such as CLIP~\cite{clip}. 
OVXD~\cite{Lin24} adapts CLIP to X-ray imagery using three learnable adapters that aggregate visual cues.
RAXO~\cite{raxo} replaces the text-based classifiers of RGB detectors with X-ray visual class descriptors built from in-house and web-retrieved samples.
X-SSL~\cite{michaelxssl} combines zero-shot proposals, multimodal clustering, and knowledge distillation for self-supervised threat localization.
Despite covering broader categories, open-vocabulary methods still require category names or visual exemplars at inference time.
In contrast, our \method enables \emph{category-agnostic object localization} in X-ray images by learning localization priors from physics-guided synthesized samples.
\textit{By removing the dependence on text prompts, web retrieval, and target-domain annotations, \method shifts the focus from semantic recognition to physics-grounded localization under X-ray superposition}.

\noindent\textbf{Foundation Models for Universal Localization.}
Promptable segmentation foundation models have reshaped generic object localization.
The Segment Anything Model (SAM)~\cite{sam1} and its variants (SAM2~\cite{sam2}, SAM3~\cite{sam3}) achieve strong zero-shot, class-agnostic segmentation by converting sparse prompts (\eg, points) into masks, benefiting from large-scale pre-training, such as on SA-1B~\cite{sam1} (11M images, 1B masks). 
Despite strong generalization on natural images, SAM becomes unreliable under extreme object superposition, weak boundaries, and low-density materials~\cite{sam_xray}.
A common solution involves domain adaptation via supervised fine-tuning or lightweight adapters~\cite{rsprompter,samadapter,polyp-sam}, but these methods require substantial labeled data, which is impractical for X-ray imagery. 
Recent works explore multi-granularity segmentation~\cite{li2024segment,zhao2024graco} and unsupervised signals for foundation models~\cite{unsamv2}.
However, their assumptions remain largely rooted in RGB imagery, where object boundaries and part-whole structures are visually separable.
\textit{This highlights the need for localization priors that are not only category-agnostic, but also aware of the physics-induced structural ambiguities of X-ray imaging}.

\noindent\textbf{Self-Supervised Segmentation and Pseudo-Labeling.}
Self-supervised visual features, especially those from ViTs, often exhibit object-aligned structures and have been widely used for label-free segmentation and proposal generation~\cite{caron2021emerging,dinov2,dinov3}. 
CutLER~\cite{cutler} leverages normalized cuts~\cite{shi2000normalized} to discover object masks, while SOHES~\cite{cao2024sohes}, UnSAM~\cite{wang2024segment}, and UnSAMv2~\cite{unsamv2} further learn multi-granularity predictions from hierarchical pseudo-labels generated through divide-and-conquer pipelines.
These methods provide useful unsupervised signals, but they largely rely on RGB-style appearance grouping and visible-surface objectness. 
Such assumptions are poorly suited to X-ray scans, where objects are projected through nonlinear attenuation and multiple items may occupy the same pixels.
\textit{Rather than discovering objects directly from arbitrary cluttered scans, \method constructs self-supervision from mined salient X-ray objects and uses a physics-guided generator to synthesize superposed training samples}.
% This design turns pseudo-labeling from an appearance-driven grouping problem into a physics-consistent localization problem tailored to X-ray imagery.

\begin{figure*}[t]
  \centering
   \includegraphics[width=\linewidth]{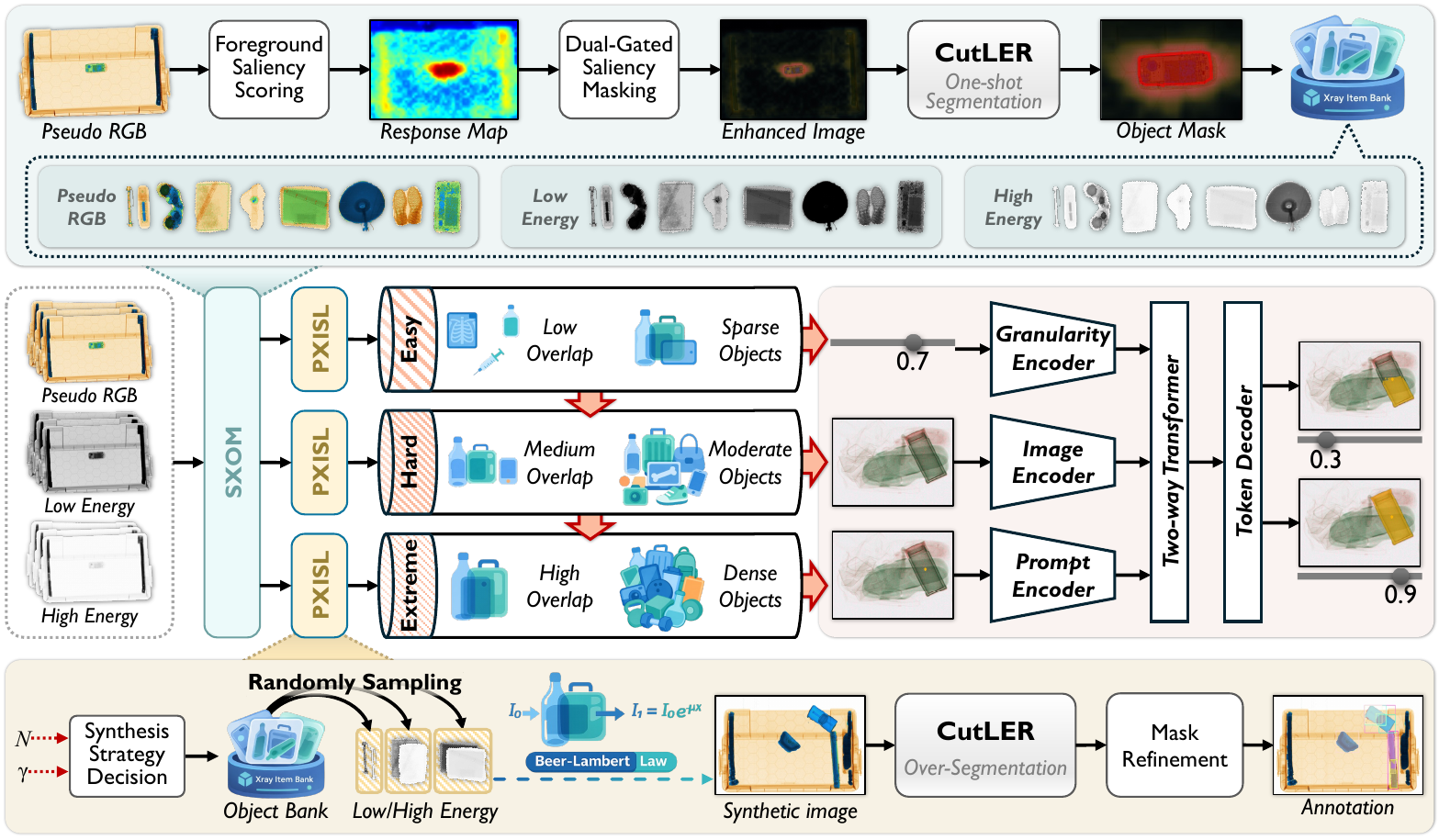}
   \caption{\textbf{Overview of \method}. Giving an unlabeled X-ray scan, we first extract foreground objects to build an item bank. Then synthesize and annotate images with randomly sampled objects to simulate realistic X-ray overlapping. Finally, we train the model with an occlusion curriculum, which enables multi-granularity localization on X-ray images without manual labels.}
   % \vspace{-15pt} % Nips submit version
   \label{fig:figure2_overview}
\end{figure*}

%%% === [Begin] New version of method ====
\section{Method}
\label{sec:methods}
As shown in Fig.~\ref{fig:figure2_overview}, \method consists of three main components.
First, we develop a \textbf{saliency-guided X-ray object mining} module to localize foreground objects from unlabeled scans by contrasting instance-specific regions against trivial backgrounds, thereby deriving an X-ray object bank of object-centric samples.
Then, we propose \textbf{physics-guided X-ray inspection synthesis and labeling} to synthesize cluttered baggage scans with physics-guided density, which been further annotated with holistic--part pseudo labels.
Finally, we introduce a \textbf{granularity supervision with occlusion-controllable training} framework to fine-tune a granularity-conditioned model with the synthesized data in an curriculum manner to enable universal object localization in realistic X-ray scenes.
Overall, \method is fully automatic and self-supervised, requiring \textit{no} manual annotations.

%% New: stage 1
\subsection{Saliency-guided X-ray Object Mining}
\label{sec:step1}
The variation of objects is the core for training models of universal object localization in X-ray scans.
However, standard unsupervised object discovery (\eg, MaskCut~\cite{cutler}) often fails to capture such objectness since faint, low-attenuation items blend into the background, while complex objects with high intensity variations appear fragmented.
Thus, following MuSc~\cite{musc}, we propose Saliency-guided X-ray Object Mining (SXOM) to separate foreground object of various density and structures via a saliency scoring mechanism, and further enhanced by Dual-Gated Saliency Masking (DGSM) (See the top of~\cref{fig:figure2_overview}).

Specifically,
let $\mathcal{T}=\{T_n\}_{n=1}^{|\mathcal{T}|}$ be an unlabeled X-ray dataset where each sample $T_n=(\mathbf{H}_n, \mathbf{L}_n, I_n)$ consists of the high-energy density map $\mathbf{H}_n$, the low-energy density map $\mathbf{L}_n$, and a pseudo-RGB image $I_n$ that obtained by mapping densities to color.
We first extract patch features $P_n = \{p_n^m\}_{m=1}^{M}$ using a frozen encoder $\phi(\cdot):\mathbbm{R}^{H\times W\times3}\rightarrow\mathbbm{R}^{d\times d \times M}$.
Then, to quantify the objectness, we define the saliency score as the average of the smallest $K = \left\lceil X \cdot (|\mathcal{T}|-1) \right\rceil$ distances between patch features:
\begin{equation}\label{eq:saliency}
s_n^m = \frac{1}{K} \sum_{i=1}^{K} \big(\text{sort}(d_{n,k}^m)_{k\neq n}\big)_i,~d_{n,k}^m = \min_{1 \le j \le M} \|p_n^m - p_k^j\|_2,
\end{equation}
where $d_{n,k}^m$ is the the nearest neighbor for patch feature $p_n^m$ in image $I_k$.

consequently, we upsample these scores into an image-level saliency map $S_n \in [0,1]^{H \times W}$, which reveal the location of foreground objects.
However, the raw saliency maps are often coarse due to internal object heterogeneity and density variation. 
Hereby, we propose a Dual-Gated Saliency Masking (DGSM) module to refine $S_n$ via two sigmoid-based gates with thresholds $\tau_1,\tau_2$:
\begin{equation}\label{eq:dual_gated}
G_{\text{sharp}} = \sigma(k~\cdot~(S_n - \tau_1 \max S_n)), \quad G_{\text{soft}} = \sigma(k~\cdot~(S_n - \tau_2 \max S_n)).
\end{equation}
$G_{\text{sharp}}$ identifies confident object cores, while $G_{\text{soft}}$ captures the broader extent. We fuse these into an adaptive weight map $W_n = \frac{1}{2}(\text{norm}(S_n^3 \odot G_{\text{soft}}) + G_{\text{sharp}})$, where the cubic term $S_n^3$ suppresses remaining background noise. We then generate a saliency-enhanced image $I_n^{\text{fuse}} = (I_n + C_n) \odot W_n$, making the foreground significantly more distinct for better mining, where $C_n \in [0,1]^{H \times W \times 3}$ denotes the JET-colorized heatmap of the saliency map $S_n$.

With the enhanced X-ray image, we can easily extract a clean instance mask $Y_n$ by adopting any off-the-shelf segmenter $\psi(\cdot)$.
Finally, we crop all aligned modalities to form an \textbf{X-ray Item Bank} $\mathcal{B} = \{\mathcal{O}_n\}^{|\mathcal{B}|}_{n=1}$, where each entry is a quadruple:
\begin{equation}
\mathcal{O}_n = (\mathbf{H}_n \odot Y_n, \mathbf{L}_n \odot Y_n, I_n \odot Y_n, Y_n) = (\hat{\mathbf{H}}_n,\hat{\mathbf{L}}_n,\hat{I}_n,Y_n).
\end{equation}
This bank $\mathcal{B}$ serves as the source for the controllable synthesis and pseudo-labeling described in Sec.~\ref{sec:step2}, which contains dual-energy density maps $(\hat{\mathbf{H}},\hat{\mathbf{L}})$ and an isolated object in pseudo-RGB image $\hat{I}$ with the corresponding mask $Y$

%% New: stage 2
\subsection{Physics-guided X-ray Inspection Synthesis and Labeling}
\label{sec:step2}
\cai{Real baggage scans are heavily cluttered: objects overlap extensively, and their appearance changes follow X-ray transmission physics rather than alpha blending. 
Moreover, manually fully annotate X-ray images is impractical due to the characteristics of X-rays, such as volume superposition, blurred boundaries, and low-attenuation targets.
Hence, we propose a Physics-guided X-ray Inspection Synthesis and Labeling (PXISL) process to simulate the real-world X-ray imagery with the X-ray item bank $\mathcal{B}$.
PXISL provides physics-guided density superposition and multi-granular holistic--part pseudo labels for both object-level and complementary fine-structure supervision to train \method.}

Given the X-ray item bank $\mathcal{B}$, we synthesize a cluttered scan by placing $N$ sampled items $\{\mathcal{O}_{i}|\mathcal{O}_{i}\in\mathcal{B}\}_{i=1}^N$ onto a real background $(\mathbf{H}^{\text{bg}},\mathbf{L}^{\text{bg}})$. 
For each sampled item, we generate a placement transform $\mathcal{J}_i$ with controllable position and random data augmentation.
Hence, the canvas-aligned maps is defined as: $(\bar{\mathbf{H}}_i,\bar{\mathbf{L}}_i,\bar{Y}_i)=\mathcal{J}_i(\hat{\mathbf{H}}_{n_i},\hat{\mathbf{L}}_{n_i},Y_{n_i})$,
where $\bar{Y}_i$ indicates the placed region on the canvas. 
To simulate the realistic X-ray scans, we project the density into the \emph{absorbance} domain following the Beer--Lambert law~\cite{swinehart1962beer}, \ie, absorbance is additive in log-transmittance for overlap. 
For each density map $e\in\{H,L\}$, we convert intensities to absorbance:
\begin{equation}
A_e^{\text{bg}}=-\log\!\left(\frac{e^{\text{bg}}}{\alpha}+\epsilon\right),
\quad
A_e^{(i)}=-\log\!\left(\frac{\bar{e}_i}{\alpha}+\epsilon\right),
\label{eq:absorbance_single_density_map}
\end{equation}
where $\bar{e}_i\in\{\bar{H}_i,\bar{L}_i\}$, $\alpha$ is the incident intensity, and $\epsilon$ is a small constant.
The overall intensities is calculated by converting the composite absorbance:
\begin{equation}
A_e^{\text{syn}} = A_e^{\text{bg}} + \sum_{i=1}^{N}\Big(\bar{Y}_i \odot A_e^{(i)}\Big),
\qquad
e^{\text{syn}}=\alpha \exp(-A_e^{\text{syn}}).
\label{eq:beer_lambert_add}
\end{equation}
Finally, we render the pseudo-RGB visualization as $I^{\text{syn}}=f_{\text{color}}(H^{\text{syn}},L^{\text{syn}})$, where $f_{\text{color}}$ is a standard density to color mapping function.
With this controllable synthesis process, we can adjust the complexity of synthesized scans via object number $N$ and the overlap ratio between objects.
Furthermore, we obtain the instance-level masks $\{\bar{Y}_i\}_{i=1}^{N}$ under this controllable synthetic process.

Nevertheless, holistic masks alone are not sufficient for X-ray overlaps, which are often \textbf{Matryoshka}-like~\cite{kusupati2022matryoshka}: a visible ``part'' can itself correspond to an entire object (or a meaningful subcomponent), so the learner must handle different mask granularities.
To obtain object-consistent part masks, we run a segment model $\overline\psi(\cdot)$ on the synthesized pseudo-RGB image $I^{\text{syn}}$ to produce an over-segmentation $\{R_j\}_{j=1}^{J}$, and filter out masks that deviated from the foreground area:
\begin{equation}
\mathcal{R}^{\text{syn}}
=
\{~R_j~|~R_j \subseteq Y^{\text{syn}}~\}, \quad \text{where}~Y^{\text{syn}}=\bigcup\nolimits_{i=1}^{N}\bar{Y}_i,
\label{eq:region_filter}
\end{equation}
which preserves local mask with natural visual cues.
Then, we design a composite granularity labeling strategy to encourage model learn the relationship between holistic objects and the components.
Following UnSAMv2~\cite{unsamv2}, we define a normalized granularity score for each part mask $R_j$:
\begin{equation}
g_j = 0.1 + 0.9\cdot\frac{\sqrt{|R_j|}-s_{\min}}{s_{\max}-s_{\min}},
\quad
s_{\max}=\max_{R\in\mathcal{R}^{\text{syn}}}\sqrt{|R|},\ 
s_{\min}=\min_{R\in\mathcal{R}^{\text{syn}}}\sqrt{|R|}.
\end{equation}
Then, we further assign $g=1$ to each holistic instance mask $\bar{Y}_i$, and remove redundant masks with highly overlap.
Overall, each synthesized sample is paired with a label pool:
\begin{equation}
\mathcal{G}^{\text{syn}}
=
\{(\bar{Y}_i,1)\}_{i=1}^{N}
\ \cup\
\{(R_j,g_j)\}_{R_j\in\mathcal{R}^{\text{syn}}}.
\end{equation}
which been further used to train the granularity-conditioned SAM2 in Sec.~\ref{sec:curriculum}.

%% New: stage 3
\subsection{Granularity Supervision with Occlusion-Controllable Training}
\label{sec:curriculum}

Now we can transfer the “localize any object” ability from RGB foundation models by fine-tuning SAM2 with the controllable synthesized scans and the corresponding holistic--part label pool.
Note that, the inherent domain gap between natural RGB and X-ray images comes from the complexity of overlapped structure and low texture visual cues.
To address this, we adopt an \textit{occlusion-controlled curriculum} and \textit{granularity-conditioned annotations}, which together stabilize adaptation with object-level and complementary fine-structure supervision from single objects to severe overlap.

\noindent\textbf{Occlusion-controlled curriculum.}
Our synthesis controls difficulty of curriculum via (i) the number of pasted items $N$ and (ii) an overlap cap $\gamma$ (maximum overlap between instance masks \cai{defined by IoU}).
We generate tiers $\{\mathcal{D}^{(t)}\}_{t=1}^{T}$ with fixed $(N^{(t)},\gamma^{(t)})$ and train from easy to hard settings (Model training in Sec.~\ref{sec:implementation details}), progressively increasing clutter and superposition.

\noindent\textbf{Granularity-conditioned SAM2 training.}
Follow UnSAMv2~\cite{unsamv2}, we augment SAM2 with a lightweight granularity encoder that injects a continuous $g\in(0,1]$ into the mask prediction pathway, enabling mask prediction at the requested granularity.
We train on synthesized scans paired with our multi-granularity label pool (holistic masks with $g{=}1$ and part masks with $g{<}1$), which teaches SAM2 to localize both whole instances and meaningful regions under \textit{Matryoshka-like} X-ray overlaps.

\section{Experiments}
\label{sec:experiments}
\subsection{Implementation Details}
\label{sec:implementation details}
\noindent\textbf{Physics-guided synthesis:}
For foreground saliency extraction (\cref{sec:step1}), we use a frozen CLIP image encoder~\cite{clip} with ViT-L/14~\cite{vit} as the feature extractor $\phi$.
For the saliency computation (\cref{eq:saliency}), we set $X=30\%$, and for the dual-gated masking (\cref{eq:dual_gated}), we set $\tau_1=0.8$, $\tau_2=0.6$, and $k=5$.
Following UnSAMv2~\cite{unsamv2}, we use CutLER~\cite{cutler} with a Cascade Mask R-CNN~\cite{Cascade_R_CNN} object detector as the off-the-shelf segment model.
We generate 18,000 synthetic training scenes using the physics-guided Beer–Lambert~\cite{swinehart1962beer} composition in \cref{sec:step2} by sampling objects uniformly from the mined item bank and applying geometric transforms (\eg, translation, scaling, rotation).

\noindent\textbf{Model training:}
We adopt a pretrained SAM2-small~\cite{sam2} as the base model for \method and adapt the granularity-conditioned SAM2 design from UnSAMv2~\cite{unsamv2} to learn from holistic and part pseudo labels.
We fine-tune the fourier encoder, its corresponding granularity mask token, and the transformer block in the mask decoder using LoRA~\cite{lora} of rank 8 while freezing the rest of the parameters.
The batch size is set to 16, and we adopt AdamW~\cite{adamw} with weight decay of 1e-5, and we use cosine scheduler to adjust learning rate during training.
All experiments are run on 8 NVIDIA H100 GPUs (80GB).
More details of occlusion-controllable curriculum learning can be found in the Appendix~\ref{app:comp}.

\noindent\textbf{Datasets:}
We use 1621 images in COMPASS-XP~\cite{compass-xp} as the X-ray scan classification dataset $\mathcal{T}$ for mining objects and synthesizing training data.
Following prior X-ray evaluation practice~\cite{raxo},
we report results on six datasets.
DET-COMPASS~\cite{raxo} and five prohibited-item datasets (\ie, CLCXray~\cite{clcxray}, DvXray~\cite{dvxray}, PIXray~\cite{pixray}, HiXray~\cite{hixray} and PIDray~\cite{pidray}).
All six datasets are test-only for \method, as
we do not use any of their images for mining, synthesis, or training.

\noindent\textbf{Evaluation Protocol:}
We evaluate category-agnostic localization with standard MS-COCO box metrics~\cite{coco}. 
For SAM-style segmenters, predicted masks are converted to tight bounding boxes.
Since several X-ray benchmarks provide only partial annotations, we restrict evaluation to predictions near annotated targets, \ie, only consider predictions which center fall inside any expanded GT boxes \cai{by \(1.5\times\)}.
COCO matching and aggregation are then applied unchanged, reporting AP$_{50}$, AP$_{75}$, mAP, and AR$_{1/10}$. 
For PIXray and PIDray, which provide pixel-level masks, we also report standard segmentation metrics, mIoU and F1, in Tab.~\ref{tab:seg_results}.
\cai{Note that, RAXO~\cite{raxo} results are adopted from the original paper under the conventional MS-COCO evaluation.}

\begin{table}[t]
  \centering
  \caption{\textbf{AP on DET-COMPASS and Five Real-World X-ray Datasets.} The best results are marked in bold, and the second-best results are underlined.
  }
  \label{tab:main_results_ap_all}
  \fontsize{10pt}{12pt}\selectfont
  \setlength{\tabcolsep}{2.5pt}
  \begin{adjustbox}{width=1\linewidth, center}
  \begin{tabular}{lcc >{\columncolor{cyan!50!green!20}}c|cc >{\columncolor{cyan!50!green!20}}c|cc >{\columncolor{cyan!50!green!20}}c|cc >{\columncolor{cyan!50!green!20}}c|cc >{\columncolor{cyan!50!green!20}}c|cc >{\columncolor{cyan!50!green!20}}c}
    \toprule
    \multirow{2}{*}{\textbf{Method}}
      & \multicolumn{3}{c|}{\textbf{DET-COMPASS}}
      & \multicolumn{3}{c|}{\textbf{CLCXray}}
      & \multicolumn{3}{c|}{\textbf{DvXray}}
      & \multicolumn{3}{c|}{\textbf{PIXray}}
      & \multicolumn{3}{c|}{\textbf{HiXray}}
      & \multicolumn{3}{c}{\textbf{PIDray}} \\
    & AP$_{50}$ & AP$_{75}$ & AP$^{95}_{50}$
    & AP$_{50}$ & AP$_{75}$ & AP$^{95}_{50}$
    & AP$_{50}$ & AP$_{75}$ & AP$^{95}_{50}$
    & AP$_{50}$ & AP$_{75}$ & AP$^{95}_{50}$
    & AP$_{50}$ & AP$_{75}$ & AP$^{95}_{50}$
    & AP$_{50}$ & AP$_{75}$ & AP$^{95}_{50}$ \\
    \midrule
    \rowcolor{gray!30} RAXO$^{100}_{0}$ & 54.20 & 48.80 & 47.90 & 29.60 & 24.40 & 22.20 & 26.60 & 24.10 & 22.60 & 45.00 & 39.00 & 36.90 & 27.20 & 19.40 & 17.10 & 21.40 & 17.90 & 16.50 \\
    \rowcolor{gray!30} RAXO$^{0}_{100}$ & 19.80 & 16.80 & 16.10 &  9.70 &  7.50 &  7.10 & 14.10 & 12.90 & 12.40 & 19.80 & 16.80 & 16.10 & 13.00 &  8.70 &  7.90 & 17.10 & 14.30 & 13.40 \\
    SAM2                                & \underline{36.10} & \underline{29.13} & \underline{29.74} & 17.75 & 14.73 & 12.80 & 10.38 &  9.72 &  8.87 &  6.33 &  5.45 &  5.14 & 18.99 & 11.22 & 10.58 &  1.97 &  1.16 &  1.13  \\
    
    SAM3                                & 11.46 &  9.93 &  9.69 & 28.14 & \underline{21.74} & 18.93 & 11.77 &  9.82 &  8.55 & 12.04 &  9.98 &  9.05 & 24.32 & 12.15 & 12.77 &  7.21 &  3.66 &  3.79 \\
    
    UnSAMv2                             & 33.25 & 27.90 & 27.44 & \underline{30.71} & 20.22 & \underline{19.06} & \underline{19.58} & \underline{15.34} & \underline{14.05} & \underline{22.54} & \underline{17.72} & \underline{16.97} & \underline{43.16} & \underline{21.14} & \underline{22.26} &  \underline{9.57} &  \underline{4.79} &  \underline{4.94}  \\

    \textbf{\method}                   & \textbf{63.39} & \textbf{55.73} & \textbf{53.33} & \textbf{37.61} & \textbf{27.44} & \textbf{25.29} & \textbf{23.56} & \textbf{18.33} & \textbf{16.87} & \textbf{30.26} & \textbf{22.40} & \textbf{21.37} & \textbf{43.35} & \textbf{21.75} & \textbf{22.47} & \textbf{11.61} & \textbf{5.16} & \textbf{5.65} \\
    \bottomrule
  \end{tabular}
  \end{adjustbox}
  % \vspace{-10pt} % Nips submit version
\end{table}

\begin{figure}[t]
    \centering
    \begin{minipage}[t]{0.69\textwidth}
    \centering
    \captionof{table}{\textbf{AR on DET-COMPASS and Five Real-World X-ray Datasets.}
                        Measuring item coverage for high confidence predictions.
                        }
    \label{tab:main_results_ar_no_det}
    \begin{adjustbox}{width=1\linewidth, center}
      \begin{tabular}{lc c|c c|c c|c c|c c|c c}
        \toprule
        \multirow{2}{*}{\textbf{Method}}
          & \multicolumn{2}{c|}{\textbf{DET-COMPASS}}
          & \multicolumn{2}{c|}{\textbf{CLCXray}}
          & \multicolumn{2}{c|}{\textbf{DvXray}}
          & \multicolumn{2}{c|}{\textbf{PIXray}}
          & \multicolumn{2}{c|}{\textbf{HiXray}}
          & \multicolumn{2}{c}{\textbf{PIDray}} \\
        & AR$_1$ & AR$_{10}$
        & AR$_1$ & AR$_{10}$
        & AR$_1$ & AR$_{10}$
        & AR$_1$ & AR$_{10}$
        & AR$_1$ & AR$_{10}$
        & AR$_1$ & AR$_{10}$ \\
        \midrule
        SAM2                & 40.49 & 47.99 & 19.52 & 29.72 & 19.60 & 25.66 &  5.36 & 21.63 &  8.91 & 28.04 &  5.94 &  8.63 \\
        
        SAM3                & 14.11 & \underline{52.10} & \underline{23.03} & \textbf{51.41} & 17.82 & \underline{34.93} &  5.01 & \underline{35.95} &  9.59 & \textbf{38.92} &  8.76 & \underline{35.68} \\
        
        UnSAMv2            & \underline{42.88} & 48.64 & 21.84 & 33.73 & \underline{25.46} & 33.66 & \textbf{14.70} & 34.83 & \textbf{17.51} & 34.84 & \textbf{13.62} & 23.34 \\
    
        \textbf{\method}   & \textbf{68.48} & \textbf{82.46} & \textbf{28.63} & \underline{50.20} & \textbf{30.54} & \textbf{46.44} & \underline{13.52} & \textbf{51.86} & \underline{16.70} & \underline{34.97} & \underline{12.54} & \textbf{37.06} \\
        \bottomrule
      \end{tabular}
      
      \end{adjustbox}
    \end{minipage}
    % \vspace{-6pt} % Nips submit version
    \hfill
    \begin{minipage}[t]{0.283\textwidth}
    \centering
    \captionof{table}{Segmentation Metrics on PIXray and PIDray.}
    \begin{adjustbox}{width=1\linewidth}
        \begin{tabular}{l| cc | cc}
          \toprule
          \multirow{2}{*}{\textbf{Method}} & \multicolumn{2}{c|}{\textbf{PIXray}} &  \multicolumn{2}{c}{\textbf{PIDray}} \\
          & mIoU & F1 & mIoU & F1 \\
          \midrule
          SAM2              & 24.68 & 28.15 & 15.15 & 19.08 \\
          SAM3              & 44.16 & 49.79 &  \underline{46.48} &  \underline{57.24} \\
          UnSAMv2           & \underline{51.52} &  \underline{60.20} & 38.24 & 48.87 \\
          \textbf{\method}  & \textbf{64.06} & \textbf{71.86} & \textbf{50.37} &\textbf{ 61.17} \\
          \bottomrule
        \end{tabular}
    \end{adjustbox}
    \label{tab:seg_results}
    \end{minipage}
    % \vspace{-6pt} % Nips submit version
\end{figure}

\begin{figure}[t]
  \centering
   \includegraphics[width=0.95\linewidth]{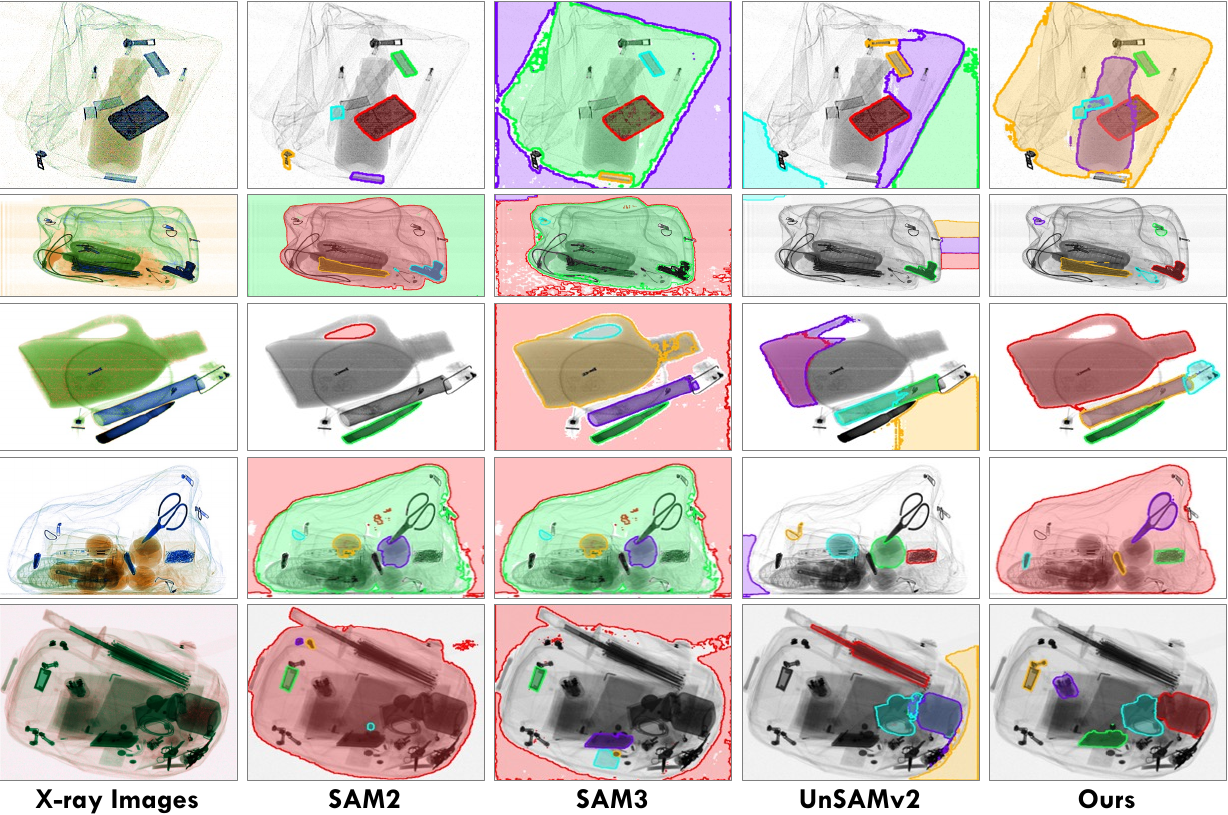}
\caption{\textbf{Comparison of masks with top-5 confidence across real-world X-ray datasets.}
To highlight the predicted mask, we convert the original image to a grayscale image in our visualizations.
\method produces more potential objects with high confident, \eg, segment the bottle in row 1.
}
   \label{fig:Mask_comparison}
\end{figure}

\subsection{Comparison with the State of the Art}
We compare \method with recent X-ray detection and universal segmentation baselines.
RAXO~\cite{raxo} is an open-vocabulary X-ray detector that builds dataset-specific class descriptors from either in-house X-ray data (RAXO$^{100}_{0}$) or Internet data (RAXO$^{0}_{100}$).
Unlike RAXO, \method uses no target category names, text queries, web retrieval, or target-domain annotations.
We also evaluate SAM2~\cite{sam2} and SAM3~\cite{sam3} as strong RGB promptable segmentation baselines.
Finally, UnSAMv2~\cite{unsamv2} is fine-tuned on the same COMPASS-XP~\cite{compass-xp} images but without our object mining, physics-guided synthesis, or occlusion curriculum, isolating the gains from \method's X-ray-specific self-supervised adaptation.

\noindent\textbf{Quantitative results.}
Tabs.~\ref{tab:main_results_ap_all} and~\ref{tab:main_results_ar_no_det} summarize bbox localization results on DET-COMPASS and five real-world X-ray benchmarks.
Overall, \method improves both localization precision and object coverage under limited predictions.
On exhaustively annotated DET-COMPASS, \method achieves the best AP$_{50}$, AP$_{75}$, and mAP, with \{63.39\%, 55.73\%, 53.33\%\}, substantially outperforming all baselines.
On the five prohibited-item datasets, \method consistently surpasses promptable segmentation baselines in AP$_{50}^{95}$, \eg increasing UnSAMv2 from 19.06\% to 25.29\% on CLCXray.
It also achieves the best $\mathrm{AR}_{10}$ on four of six benchmarks, \eg improving over SAM3 from 34.93\% to 46.44\% on DvXray.
These results show that \method effectively adapts SAM-style localization to X-ray superposition through physics-guided synthesis and occlusion-controlled fine-tuning.
RAXO remains a strong semantic detector on prohibited-item benchmarks, while \method surpasses it on DET-COMPASS and achieves the best AP on CLCXray and HiXray, highlighting complementary advantages beyond vocabulary-driven detection.

For PIXray and PIDray, which provide pixel-level masks, we further report standard \textbf{segmentation metrics} in Tab.~\ref{tab:seg_results}.
\method achieves the best mIoU and F1 on both datasets, \eg improving over the strongest baseline by more than 10\% on PIXray, reaching 64.06\%/71.86\%.
This confirms that \method learns more precise object structure from curriculum training on synthesized scans.

\noindent\textbf{Qualitative comparison.}
Fig.~\ref{fig:Mask_comparison} visualizes the top-5 confident predictions. 
RGB foundation models show a clear X-ray domain gap: SAM2 misses overlapped or low-density objects, such as bottles in the first and third rows, while SAM3 often produces oversized noisy regions spilling into the background.
UnSAMv2 improves X-ray-oriented localization but still struggles with heavy superposition, causing unstable extents and incomplete instance separation.
In contrast, \method produces more object-aligned high-confident masks, preserving fine structures and separating overlapping items more consistently.
Notably, in the third row, \method accurately segments the plastic jug, bullet, and air pump, including the foot peg and valve head.
These items are not explicitly boxed by manual annotations, suggesting stronger class-agnostic localization beyond labeled targets.

\begin{table}[t]
  \centering
  \caption{Ablation study for our proposed curriculum training strategy for different combinations of synthetic scenes (Easy/Hard/Extreme) and combination of all samples used for curriculum learning.}
  \begingroup
  \fontsize{8pt}{10pt}\selectfont
  \setlength{\tabcolsep}{2pt}
  \setlength{\mathsurround}{2pt}
  \renewcommand{\arraystretch}{1.2}
  \begin{adjustbox}{width=1\linewidth, center}
    \begin{tabular}{@{}c|cccc|ccccc|ccccc|ccccc|ccccc|ccccc|ccccc@{}}
      \toprule
      & \multicolumn{4}{c|}{\textbf{}} & \multicolumn{5}{c|}{\textbf{CLCXray}} & \multicolumn{5}{c|}{\textbf{DvXray}} &  \multicolumn{5}{c|}{\textbf{PIXray}} & \multicolumn{5}{c|}{\textbf{HiXray}} & \multicolumn{5}{c|}{\textbf{PIDray}} & \multicolumn{5}{c}{\textbf{Average}}\\
      & Easy & Hard & Extreme & Curriculum & AP$_{50}$ & AP$_{75}$ & AP$^{95}_{50}$ & AR$_1$ & AR$_{10}$ 
      & AP$_{50}$ & AP$_{75}$ & AP$^{95}_{50}$ & AR$_1$ & AR$_{10}$ 
      & AP$_{50}$ & AP$_{75}$ & AP$^{95}_{50}$ & AR$_1$ & AR$_{10}$ 
      & AP$_{50}$ & AP$_{75}$ & AP$^{95}_{50}$ & AR$_1$ & AR$_{10}$ 
      & AP$_{50}$ & AP$_{75}$ & AP$^{95}_{50}$ & AR$_1$ & AR$_{10}$
      & AP$_{50}$ & AP$_{75}$ & AP$^{95}_{50}$ & AR$_1$ & AR$_{10}$ \\
      \midrule
      (1) & \checkmark & \checkmark & \checkmark &                    &
      \textbf{38.41} & \textbf{28.47} & \underline{25.21} & \underline{27.82} & \underline{40.62} 
      & 19.99 & 14.73 & 13.69 & 25.63 & \underline{41.81}
      & 24.15 & 16.57 & 16.13 &  9.01 & \underline{46.77} 
      & 42.86 & 22.02 & 22.33 & \underline{17.94} & 33.73 
      & \underline{10.28} &  \underline{4.70} &  \underline{5.09} & \underline{11.68} & \underline{36.18} 
      & 27.14 & 17.30 & 16.49 & 18.41 & 39.82 \\	
      (2) & \checkmark & & & \checkmark                     &
      35.29 & 24.84 & 22.92 & 26.84 & 38.92 
      & 16.69 & 12.90 & 11.66 & 22.96 & 35.09 
      & 21.17 & 15.27 & 14.75 &  8.00 & 40.51 
      & \textbf{45.07} & \textbf{23.42} & \textbf{23.65} & \textbf{18.46} & \textbf{35.11} 
      &  8.30 &  2.90 &  3.58 &  9.65 & 27.38 
      & 25.76 & 16.11 & 15.58 & 17.54 & 34.90 \\ 
      (3) & \checkmark & \checkmark & & \checkmark             &
      36.85 & 27.07 & 24.16 & 27.62 & 36.83 
      & \underline{20.86} & \underline{15.56} & \underline{14.46} & \underline{26.89} & 41.19 
      & \underline{25.93} & \underline{18.48} & \underline{17.80} & \underline{10.81} & 46.70 
      & \underline{44.39} & \underline{22.29} & \underline{23.08} & 17.04 & 34.91 
      &  9.58 &  3.94 &  4.47 & 11.09 & 32.90 
      & 27.61 & 17.48 & 16.83 & 18.75 & 36.45	\\
      (4) & \checkmark & \checkmark & \checkmark & \checkmark
      & \underline{37.61} & \underline{27.44} & \textbf{25.29} & \textbf{28.63} & \textbf{50.20}
      & \textbf{23.56} & \textbf{18.33} & \textbf{16.87} & \textbf{30.54} & \textbf{46.44} 
      & \textbf{30.26} & \textbf{22.40} & \textbf{21.37} & \textbf{13.52} & \textbf{51.86} 
      & 43.35 & 21.75 & 22.47 & 16.70 & \underline{34.97} 
      & \textbf{11.61} & \textbf{5.16} & \textbf{5.65} & \textbf{12.54} & \textbf{37.06} 
      & \textbf{29.28} & \textbf{19.01} & \textbf{18.33} & \textbf{20.38} & \textbf{44.11} \\
      \bottomrule
    \end{tabular}
  \end{adjustbox}
  \endgroup
\label{ablation}
\end{table}

\begin{figure}[t]
  \centering
   \includegraphics[width=0.95\linewidth]{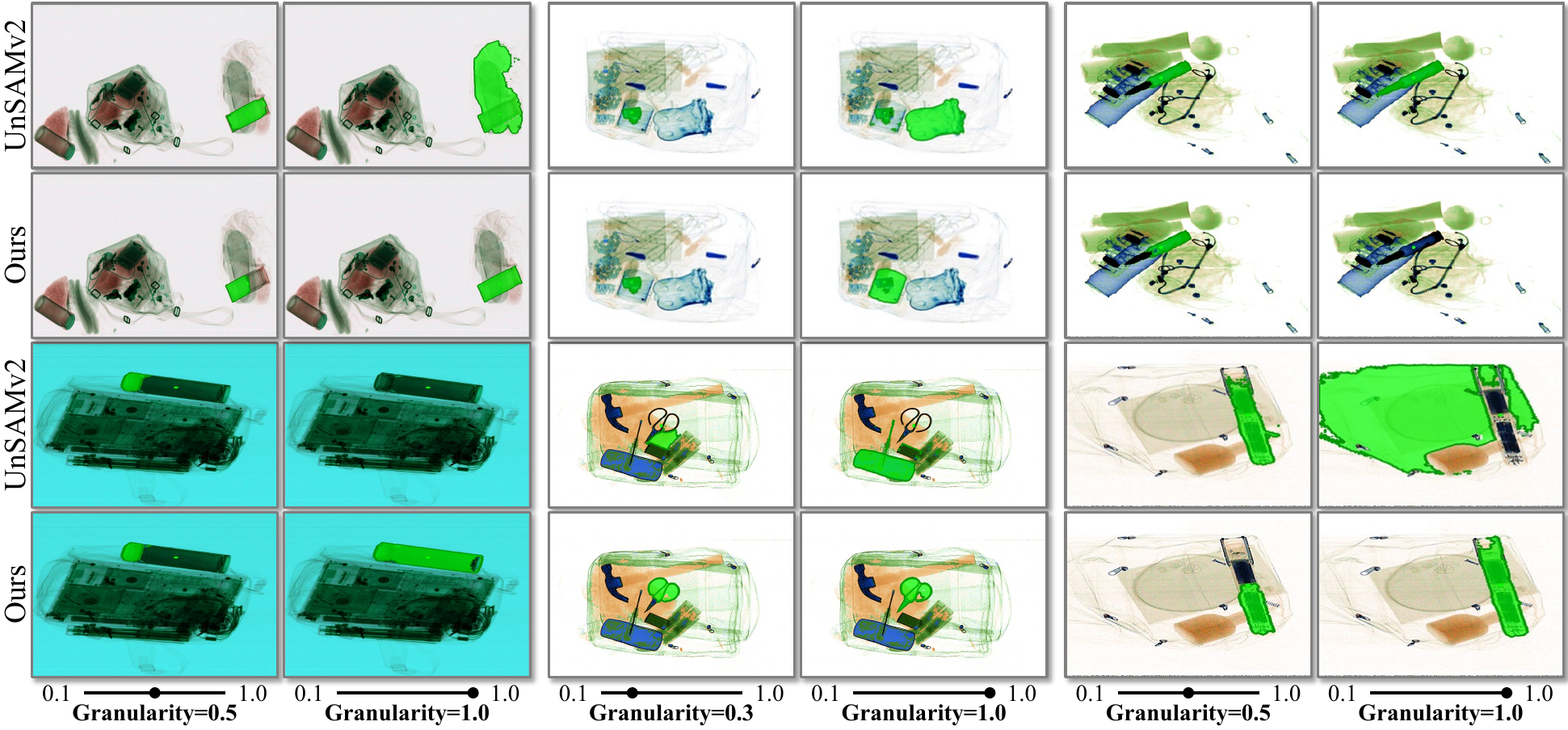}
   \caption{\textbf{Granularity comparison with UnSAMv2.} Green overlays are predicted masks at the queried granularity $g$ (shown under each column; same image prompt, only $g$ changes).
   UnSAMv2 is unstable and may collapse to over-segmenting large regions, while \method follows $g$ more consistently, transitioning from part-level (smaller $g$) to holistic masks (larger $g$) under heavy overlap.}
   \label{fig:Interactive_comparison}
   
    \centering
    \begin{minipage}[t]{0.45\textwidth}
    \centering
        \includegraphics[width=0.8\linewidth]{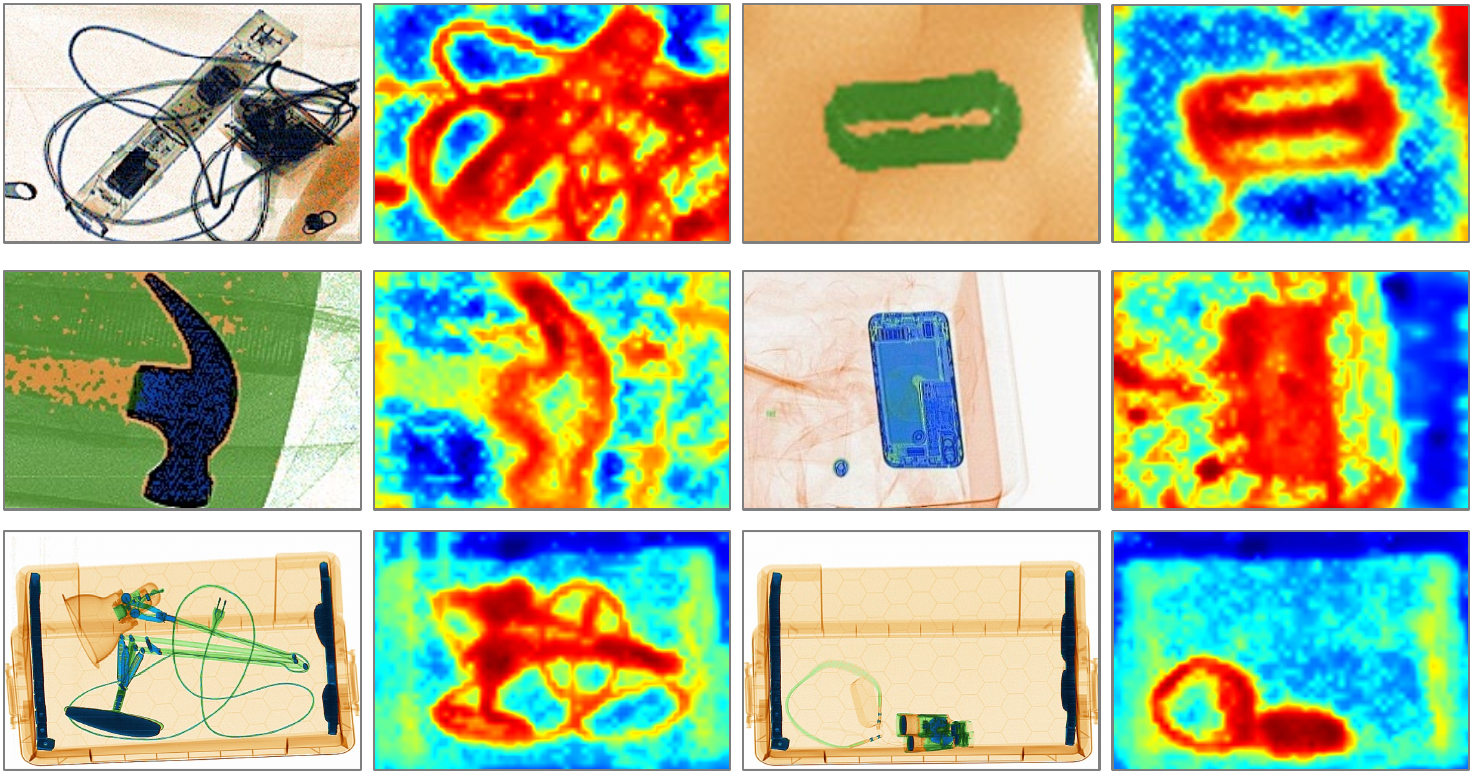}
    \captionof{figure}{Visualization of Saliency Response.}
    \label{fig:saliency_response}
    \end{minipage}
    \hfill
    \begin{minipage}[t]{0.53\textwidth}
    \centering
        \includegraphics[width=0.8\linewidth]{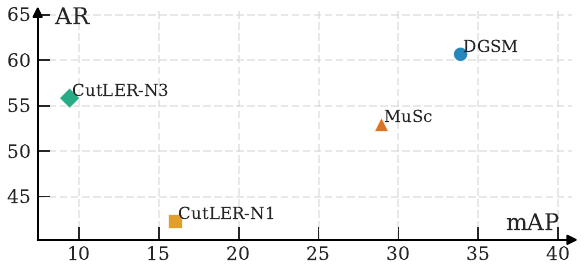}
        \captionof{figure}{Object bank construction comparison.}
        \label{fig:object_bank_comparison}
    \end{minipage}
\end{figure}

\subsection{Ablation Study}
\noindent\textbf{Occlusion-controllable curriculum:}
To isolate occlusion-controllable training, we compare \method trained on all synthetic scenes at once with the stage-wise Easy $\rightarrow$ Hard $\rightarrow$ Extreme schedule.
Tab.~\ref{ablation} shows that curriculum training improves localization across all metrics.
Comparing Rows (1) and (4), \method raises mAP from 16.49\% to 18.33\% over direct training on mixed synthetic samples, indicating that mixed-complexity scenes are less effective for adapting RGB-pretrained SAM-style models to X-rays.
Rows (2)--(4) further show consistent gains as occlusion increases gradually, with mAP rising from 15.58\% to 16.83\% and finally to 18.33\%.
This confirms that the object number $N$ and occlusion level $\gamma$ serve as effective curriculum-difficulty measures.
Overall, progressively increasing occlusion provides a smoother adaptation path, stabilizing optimization from isolated objects to densely overlapped baggage scans and improving recall under X-ray superposition.

\noindent\textbf{Interactive segmentation comparison:}
We study the capabilities of \method for interactive segmentation by querying images with the same point prompt yet different granularity values $g$.
As shown in Fig.~\ref{fig:Interactive_comparison},
UnSAMv2 exhibits brittle granularity behavior:
changing $g$ can trigger abrupt and often incorrect mask shifts (\eg, expanding to background/baggage regions) or fail to produce meaningful part-level masks under overlap.
In contrast, \method responds to $g$ in a stable and monotonic manner: 
smaller $g$ yields compact part-like regions,
while a larger $g$ progressively expands toward holistic object masks,
all while preserving plausible boundaries and instance separation in cluttered scans.
This qualitative study highlights the benefit of our holistic-part supervision and occlusion-controllable curriculum for learning reliable granularity-conditioned localization in X-ray.

\noindent\textbf{Additional analysis of saliency mining.}
Fig.~\ref{fig:saliency_response} visualizes the saliency response across various scenes, showing that our scoring mechanism can highlight foreground items with diverse densities, especially for weak attenuation, and complex internal structures. 
Although such responses are effective across different X-ray scenes, we construct the object bank from COMPASS-XP as it contains richer object categories together with dual-energy scans, offering more diverse density patterns for physics-guided synthesis. 
Fig.~\ref{fig:object_bank_comparison} further compares object mining quality on DET-COMPASS. 
Except for CutLER-N3, which predicts 3 masks, each method produces one mask per image.
LAO-X achieves the best mAP and AR, outperforming direct CutLER and MuSc, which indicates that DGSM improves the accuracy and completeness of mined object masks, providing a cleaner object bank for subsequent synthesis.

\section{Conclusion}
In this paper, we present \method, a self-supervised framework for \emph{Localizing Any Object} in X-ray security scans without manual annotations. 
\method first mines object-centric items from unlabeled scans via saliency-guided object extraction, recomposes them with physic-guided synthesis to trains a granularity-conditioned SAM2 with an occlusion-controllable curriculum. 
This pipeline turns unlabeled X-ray data into structured supervision and enables reliable localization of both whole and parts of objects under severe superposition. 
Experiments on six X-ray benchmarks show consistent improvements over RGB foundation models and self-supervised baselines in precision, recall, and mask quality. 
One limitation is the domain gap caused by pseudo-color mapping of different scanner, where we can learn a color-mapping sampling model in the future work to enhance synthesis process to improve robustness across scanners.
We hope \method can serve as a strong and practical baseline for universal X-ray perception and stimulate future work on scalable X-ray localization.

\bibliographystyle{abbrvnat}
\bibliography{main}

@String(PAMI = {IEEE Transactions on Pattern Analysis and Machine Intelligence})

@String(CVPR = {IEEE/CVF Conference on Computer Vision and Pattern Recognition})

@String(ICCV = {IEEE/CVF International Conference on Computer Vision})

@String(ECCV = {European Conference on Computer Vision})

@String(NeurIPS = {Advances in Neural Information Processing Systems})

@String(ICML = {International Conference on Machine Learning})

@String(ICLR = {International Conference on Learning Representations})

@String(CVPRW = {IEEE/CVF Conference on Computer Vision and Pattern Recognition Workshops})

@String(ITNEC = {IEEE Information Technology, Networking, Electronic and Automation Control Conference})

@String(TMM = {IEEE Transactions on Multimedia})

@String(TIFS = {IEEE Transactions on Information Forensics and Security})

@inproceedings{sam1,
  title={Segment anything},
  author={Kirillov, Alexander and Mintun, Eric and Ravi, Nikhila and Mao, Hanzi and Rolland, Chloe and Gustafson, Laura and Xiao, Tete and Whitehead, Spencer and Berg, Alexander C and Lo, Wan-Yen and others},
  booktitle=ICCV,
  pages={4015--4026},
  year={2023}
}

@inproceedings{sam2,
  title={Sam 2: Segment anything in images and videos},
  author={Ravi, Nikhila and Gabeur, Valentin and Hu, Yuan-Ting and Hu, Ronghang and Ryali, Chaitanya and Ma, Tengyu and Khedr, Haitham and R{\"a}dle, Roman and Rolland, Chloe and Gustafson, Laura and others},
  booktitle=ICLR,
  volume={2025},
  pages={28085--28128},
  year={2025}
}

@inproceedings{sam3,
  title={Sam 3: Segment anything with concepts},
  author={Carion, Nicolas and Gustafson, Laura and Hu, Yuan-Ting and Debnath, Shoubhik and Hu, Ronghang and Suris Coll-Vinent, Didac and Ryali, Chaitanya and Alwala, Kalyan Vasudev and Khedr, Haitham and Huang, Andrew and others},
  booktitle=ICLR,
  volume={2026},
  pages={138846--138923},
  year={2026}
}

@article{unsamv2,
  title={UnSAMv2: Self-Supervised Learning Enables Segment Anything at Any Granularity},
  author={Yu, Junwei and Darrell, Trevor and Wang, XuDong},
  journal={arXiv preprint arXiv:2511.13714},
  year={2025}
}

@inproceedings{cutler,
  title={Cut and learn for unsupervised object detection and instance segmentation},
  author={Wang, Xudong and Girdhar, Rohit and Yu, Stella X and Misra, Ishan},
  booktitle=CVPR,
  pages={3124--3134},
  year={2023}
}

@inproceedings{raxo,
  title={Superpowering open-vocabulary object detectors for x-ray vision},
  author={Garcia-Fernandez, Pablo and Vaquero, Lorenzo and Liu, Mingxuan and Xue, Feng and Cores, Daniel and Sebe, Nicu and Mucientes, Manuel and Ricci, Elisa},
  booktitle=ICCV,
  pages={20770--20779},
  year={2025}
}

@article{pixray,
  title={Automated segmentation of prohibited items in X-ray baggage images using dense de-overlap attention snake},
  author={Ma, Bowen and Jia, Tong and Su, Min and Jia, Xiaodong and Chen, Dongyue and Zhang, Yichun},
  journal=TMM,
  volume={25},
  pages={4374--4386},
  year={2022},
  publisher={IEEE}
}

@article{dvxray,
  title={Toward dual-view X-ray baggage inspection: A large-scale benchmark and adaptive hierarchical cross refinement for prohibited item discovery},
  author={Ma, Bowen and Jia, Tong and Li, Mingyuan and Wu, Songsheng and Wang, Hao and Chen, Dongyue},
  journal=TIFS,
  volume={19},
  pages={3866--3878},
  year={2024},
  publisher={IEEE}
}

@inproceedings{hixray,
  title={Towards real-world X-ray security inspection: A high-quality benchmark and lateral inhibition module for prohibited items detection},
  author={Tao, Renshuai and Wei, Yanlu and Jiang, Xiangjian and Li, Hainan and Qin, Haotong and Wang, Jiakai and Ma, Yuqing and Zhang, Libo and Liu, Xianglong},
  booktitle=ICCV,
  pages={10923--10932},
  year={2021}
}

@inproceedings{pidray,
  title={Towards real-world prohibited item detection: A large-scale x-ray benchmark},
  author={Wang, Boying and Zhang, Libo and Wen, Longyin and Liu, Xianglong and Wu, Yanjun},
  booktitle=ICCV,
  pages={5412--5421},
  year={2021}
}

@article{kusupati2022matryoshka,
  title={Matryoshka representation learning},
  author={Kusupati, Aditya and Bhatt, Gantavya and Rege, Aniket and Wallingford, Matthew and Sinha, Aditya and Ramanujan, Vivek and Howard-Snyder, William and Chen, Kaifeng and Kakade, Sham and Jain, Prateek and others},
  journal=NeurIPS,
  volume={35},
  pages={30233--30249},
  year={2022}
}

@article{swinehart1962beer,
  title={The beer-lambert law},
  author={Swinehart, Donald F},
  journal={Journal of chemical education},
  volume={39},
  number={7},
  pages={333},
  year={1962},
  publisher={ACS Publications}
}

@article{clcxray,
  title={Detecting overlapped objects in X-ray security imagery by a label-aware mechanism},
  author={Zhao, Cairong and Zhu, Liang and Dou, Shuguang and Deng, Weihong and Wang, Liang},
  journal=TIFS,
  volume={17},
  pages={998--1009},
  year={2022},
  publisher={IEEE}
}

@inproceedings{clip,
  title={Learning transferable visual models from natural language supervision},
  author={Radford, Alec and Kim, Jong Wook and Hallacy, Chris and Ramesh, Aditya and Goh, Gabriel and Agarwal, Sandhini and Sastry, Girish and Askell, Amanda and Mishkin, Pamela and Clark, Jack and others},
  booktitle=ICML,
  pages={8748--8763},
  year={2021},
  organization={PmLR}
}

@inproceedings{musc,
  title={Musc: Zero-shot industrial anomaly classification and segmentation with mutual scoring of the unlabeled images},
  author={Li, Xurui and Huang, Ziming and Xue, Feng and Zhou, Yu},
  booktitle=ICLR,
  volume={2024},
  pages={2172--2197},
  year={2024}
}

@article{dinov3,
  title={Dinov3},
  author={Sim{\'e}oni, Oriane and Vo, Huy V and Seitzer, Maximilian and Baldassarre, Federico and Oquab, Maxime and Jose, Cijo and Khalidov, Vasil and Szafraniec, Marc and Yi, Seungeun and Ramamonjisoa, Micha{\"e}l and others},
  journal={arXiv preprint arXiv:2508.10104},
  year={2025}
}

@article{dinov2,
  title={Dinov2: Learning robust visual features without supervision},
  author={Oquab, Maxime and Darcet, Timoth{\'e}e and Moutakanni, Th{\'e}o and Vo, Huy and Szafraniec, Marc and Khalidov, Vasil and Fernandez, Pierre and Haziza, Daniel and Massa, Francisco and El-Nouby, Alaaeldin and others},
  journal={arXiv preprint arXiv:2304.07193},
  year={2023}
}

@article{lora,
  title={Lora: Low-rank adaptation of large language models.},
  author={Hu, Edward J and Shen, Yelong and Wallis, Phillip and Allen-Zhu, Zeyuan and Li, Yuanzhi and Wang, Shean and Wang, Liang and Chen, Weizhu and others},
  journal=ICLR,
  volume={1},
  number={2},
  pages={3},
  year={2022}
}

@article{xray_detction1_physical_scale,
  title={Detecting prohibited objects with physical size constraint from cluttered X-ray baggage images},
  author={Chang, An and Zhang, Yu and Zhang, Shunli and Zhong, Leisheng and Zhang, Li},
  journal={Knowledge-Based Systems},
  volume={237},
  pages={107916},
  year={2022},
  publisher={Elsevier}
}

@article{xray_detection2_antioverlap,
  title={Ao-detr: Anti-overlapping detr for x-ray prohibited items detection},
  author={Li, Mingyuan and Jia, Tong and Wang, Hao and Ma, Bowen and Lu, Hui and Lin, Shuyang and Cai, Da and Chen, Dongyue},
  journal={IEEE Transactions on Neural Networks and Learning Systems},
  volume={36},
  number={7},
  pages={12076--12090},
  year={2024},
  publisher={IEEE}
}

@article{compass-xp,
  title={Compass-xp dataset},
  author={Griffin, LD and Caldwell, M and Andrews, JTA},
  journal={Computational Security Science Group, UCL},
  pages={4700--4708},
  year={2019}
}

@inproceedings{eds,
  title={Exploring endogenous shift for cross-domain detection: A large-scale benchmark and perturbation suppression network},
  author={Tao, Renshuai and Li, Hainan and Wang, Tianbo and Wei, Yanlu and Ding, Yifu and Jin, Bowei and Zhi, Hongping and Liu, Xianglong and Liu, Aishan},
  booktitle=CVPR,
  pages={21157--21167},
  year={2022},
  organization={IEEE}
}

@inproceedings{10082660,
  title={Optimization and research of suspicious object detection algorithm in x-ray image},
  author={Wang, Ruxue and Shi, Yuliang and Cai, Mingyu},
  booktitle={2023 IEEE 6th Information Technology, Networking, Electronic and Automation Control Conference (ITNEC)},
  volume={6},
  pages={1357--1361},
  year={2023},
  organization={IEEE}
}

@article{sima2024multi,
  title={Multi-Scale Feature Attention-DEtection TRansformer: Multi-Scale Feature Attention for security check object detection},
  author={Sima, Haifeng and Chen, Bailiang and Tang, Chaosheng and Zhang, Yudong and Sun, Junding},
  journal={IET Computer Vision},
  volume={18},
  number={5},
  pages={613--625},
  year={2024},
  publisher={Wiley Online Library}
}

@article{Lin24,
  title={Detection of novel prohibited item categories for real-world security inspection},
  author={Lin, Shuyang and Jia, Tong and Wang, Hao and Ma, Bowen and Li, Mingyuan and Chen, Dongyue},
  journal={Engineering Applications of Artificial Intelligence},
  volume={144},
  pages={110110},
  year={2025},
  publisher={Elsevier}
}

@inproceedings{li2024segment,
  title={Segment and recognize anything at any granularity},
  author={Li, Feng and Zhang, Hao and Sun, Peize and Zou, Xueyan and Liu, Shilong and Li, Chunyuan and Yang, Jianwei and Zhang, Lei and Gao, Jianfeng},
  booktitle=ECCV,
  pages={467--484},
  year={2024},
  organization={Springer}
}

@inproceedings{zhao2024graco,
  title={Graco: Granularity-controllable interactive segmentation},
  author={Zhao, Yian and Li, Kehan and Cheng, Zesen and Qiao, Pengchong and Zheng, Xiawu and Ji, Rongrong and Liu, Chang and Yuan, Li and Chen, Jie},
  booktitle=CVPR,
  pages={3501--3510},
  year={2024}
}

@inproceedings{samadapter,
  title={Sam-adapter: Adapting segment anything in underperformed scenes},
  author={Chen, Tianrun and Zhu, Lanyun and Deng, Chaotao and Cao, Runlong and Wang, Yan and Zhang, Shangzhan and Li, Zejian and Sun, Lingyun and Zang, Ying and Mao, Papa},
  booktitle=ICCV,
  pages={3367--3375},
  year={2023}
}

@article{rsprompter,
  title={RSPrompter: Learning to prompt for remote sensing instance segmentation based on visual foundation model},
  author={Chen, Keyan and Liu, Chenyang and Chen, Hao and Zhang, Haotian and Li, Wenyuan and Zou, Zhengxia and Shi, Zhenwei},
  journal={IEEE Transactions on Geoscience and Remote Sensing},
  year={2024},
  publisher={IEEE}
}

@inproceedings{polyp-sam,
  title={Polyp-sam: Transfer sam for polyp segmentation},
  author={Li, Yuheng and Hu, Mingzhe and Yang, Xiaofeng},
  booktitle={Medical Imaging 2024: Computer-Aided Diagnosis},
  volume={12927},
  pages={759--765},
  year={2024},
  organization={SPIE}
}

@article{michaelxssl,
  title={X-SSL: Self-supervised X-ray threat detection with zero-shot and multi-modal learning},
  author={Michael, Yonathan and Alansari, Mohamad and Ahmed, Abdelfatah and Werghi, Naoufel and Henschel, Andreas},
  journal={Information Processing \& Management},
  volume={63},
  number={1},
  pages={104339},
  year={2026},
  publisher={Elsevier}
}

@article{shi2000normalized,
  title={Normalized cuts and image segmentation},
  author={Shi, Jianbo and Malik, Jitendra},
  journal=PAMI,
  volume={22},
  number={8},
  pages={888--905},
  year={2000}
}

@inproceedings{cao2024sohes,
  title={SOHES: Self-supervised open-world hierarchical entity segmentation},
  author={Cao, Shengcao and Gu, Jiuxiang and Kuen, Jason and Tan, Hao and Zhang, Ruiyi and Zhao, Handong and Nenkova, Ani and Gui, Liangyan and Sun, Tong and Wang, Yu-Xiong},
  booktitle=ICLR,
  volume={2024},
  pages={16717--16738},
  year={2024}
}

@article{wang2024segment,
  title={Segment anything without supervision},
  author={Wang, XuDong and Yang, Jingfeng and Darrell, Trevor},
  journal=NeurIPS,
  volume={37},
  pages={138731--138755},
  year={2024}
}

@inproceedings{caron2021emerging,
  title={Emerging properties in self-supervised vision transformers},
  author={Caron, Mathilde and Touvron, Hugo and Misra, Ishan and J{\'e}gou, Herv{\'e} and Mairal, Julien and Bojanowski, Piotr and Joulin, Armand},
  booktitle=ICCV,
  pages={9650--9660},
  year={2021}
}

@inproceedings{curriculum_1_weakSam,
  title={Curriculum point prompting for weakly-supervised referring image segmentation},
  author={Dai, Qiyuan and Yang, Sibei},
  booktitle=CVPR,
  pages={13711--13722},
  year={2024}
}

@article{Cascade_R_CNN,
  title={Cascade R-CNN: High quality object detection and instance segmentation},
  author={Cai, Zhaowei and Vasconcelos, Nuno},
  journal=PAMI,
  volume={43},
  number={5},
  pages={1483--1498},
  year={2019},
  publisher={IEEE}
}

@inproceedings{coco,
  title={Microsoft coco: Common objects in context},
  author={Lin, Tsung-Yi and Maire, Michael and Belongie, Serge and Hays, James and Perona, Pietro and Ramanan, Deva and Doll{\'a}r, Piotr and Zitnick, C Lawrence},
  booktitle=ECCV,
  pages={740--755},
  year={2014},
  organization={Springer}
}

@article{adamw,
  title={Decoupled weight decay regularization},
  author={Loshchilov, Ilya and Hutter, Frank},
  journal={arXiv preprint arXiv:1711.05101},
  year={2017}
}

@article{vit,
  title={An image is worth 16x16 words: Transformers for image recognition at scale},
  author={Dosovitskiy, Alexey and Beyer, Lucas and Kolesnikov, Alexander and Weissenborn, Dirk and Zhai, Xiaohua and Unterthiner, Thomas and Dehghani, Mostafa and Minderer, Matthias and Heigold, Georg and Gelly, Sylvain and others},
  journal={arXiv preprint arXiv:2010.11929},
  year={2020}
}

@inproceedings{sam_xray,
  title={Performance evaluation of segment anything model with variational prompting for application to non-visible spectrum imagery},
  author={Gaus, Yona Falinie A and Bhowmik, Neelanjan and Isaac-Medina, Brian KS and Breckon, Toby P},
  booktitle=CVPRW,
  pages={3142--3152},
  year={2024}
}

@book{3d_print,
  title={3D printing and international security: risks and challenges of an emerging technology},
  author={Fey, Marco},
  volume={144},
  year={2017},
  publisher={DEU}
}

@article{Singh24,
  title={Advancements in machine learning techniques for threat item detection in X-ray images: a comprehensive survey},
  author={Singh, Archana and Dhiraj},
  journal={International Journal of Multimedia Information Retrieval},
  volume={13},
  number={4},
  pages={40},
  year={2024},
  publisher={Springer}
}

@article{rebuffel2007dual,
  title={Dual-energy X-ray imaging: benefits and limits},
  author={Rebuffel, V{\'e}ronique and Dinten, Jean-Marc},
  journal={Insight-non-destructive testing and condition monitoring},
  volume={49},
  number={10},
  pages={589--594},
  year={2007},
  publisher={The British Institute of Non-Destructive Testing}
}

@inproceedings{macdonald2001design,
  title={Design and implementation of a dual-energy X-ray imaging system for organic material detection in an airport security application},
  author={Macdonald, Richard DR},
  booktitle={Machine Vision Applications in Industrial Inspection IX},
  volume={4301},
  pages={31--41},
  year={2001},
  organization={SPIE}
}

@article{khan2020review,
  title={A review of airport dual energy X-ray baggage inspection techniques: Image enhancement and noise reduction},
  author={Khan, Sajid Ullah and Khan, Imran Ullah and Ullah, Imdad and Saif, Naveed and Ullah, Irfan},
  journal={Journal of X-ray Science and Technology},
  volume={28},
  number={3},
  pages={481--505},
  year={2020},
  publisher={SAGE Publications Sage UK: London, England}
}

@article{benedykciuk2021material,
  title={Material classification in X-ray images based on multi-scale CNN},
  author={Benedykciuk, Emil and Denkowski, Marcin and Dmitruk, Krzysztof},
  journal={Signal, Image and Video Processing},
  volume={15},
  number={6},
  pages={1285--1293},
  year={2021},
  publisher={Springer}
}

%%%%%%%%%%%%%%%%%%%%%%%%%%%%%%%%%%%%%%%%%%%%%%%%%%%%%%%%%%%%
\appendix
%%%%%%%%%%%%%%%%%%%%
% SUPPLEMENTARY MATERIALS
%%%%%%%%%%%%%%%%%%%%
\clearpage
\maketitle
\title{Supplementary Materials - Localize Any Object in X-Ray Security Scans} 
\setcounter{page}{1}

\section*{Overview}
In this appendix, we provide additional descriptions of the following contents:
\begin{itemize}
\item Implementation Details (\cref{app:implement}).
\item Partial-Annotation Evaluation Protocol and Sensitivity Analysis (\cref{app:protocol}).
\item Segmentation evaluation on real X-ray inspection dataset (\cref{app:seg}).
\item Pseudo label of holistic/partial objects and their combination (\cref{app:pseudo}).
\item Synthetic Scene Complexity and Data Scale (\cref{app:comp}).
\item Qualitative comparison of mask localization performance (\cref{app:mask_crop}).
\item Quantitative and Qualitative analysis of interactive mask prediction across multi-granularity (\cref{app:interactive}).
\item Failure Cases and Limitations (\cref{app:failure}).
\item Broader Impact discussion (\cref{app:broaderimpact}).
\end{itemize}

\section{More on Implementation Details}
\label{app:implement}
\noindent\textbf{Implementation Details of Promptable Segmentation Baselines.}
We evaluate SAM2~\cite{sam2} and SAM3~\cite{sam3} in the automatic mask generation setting for zero-shot X-ray promptable instance segmentation. 
Meanwhile, we calibrate their mask thresholds on each dataset's validation split and reporting the best validated results for fairness.
No ground-truth boxes, points, masks, or category texts are used as prompts. Instead, each model internally samples a regular grid of point prompts over the input image and generates a set of candidate masks, which are then filtered by the model's predicted mask quality, stability score, binarization threshold, and non-maximum suppression.

For SAM2, we use the SAM2.1 small checkpoint with the Ultralytics SAM interface. Images are resized to a square resolution of \(1024\times1024\), and automatic mask generation uses the default point grid. For SAM3, we use the official image model with the HuggingFace mask-generation pipeline. Images are processed in RGB format, and the automatic mask generation pipeline samples a denser point grid. We use point-prompt automatic generation only, without text prompts, ground-truth annotations, or multi-scale crop inference.

Since mask binarization and confidence filtering can strongly affect automatic mask generation results, we calibrate the relevant thresholds on the validation split of each dataset while keeping the remaining generation settings fixed. All predictions are saved in COCO instance segmentation format, and predicted masks are converted to tight bounding boxes for box-level evaluation. For class-agnostic evaluation, all predicted instances are assigned to a single foreground category.

\noindent\textbf{Dual-energy X-ray imaging.}
X-ray security scanners can be implemented with different acquisition modes, each providing a distinct trade-off between deployability, geometric coverage, and material awareness. 
Single-energy transmission imaging records only line-integrated attenuation, and is therefore simple and widely used, but it entangles material thickness, density, and composition in a single response. 
Dual-energy X-ray imaging mitigates this ambiguity by acquiring low-energy and high-energy measurements, whose energy-dependent attenuation differences provide material-related cues such as effective atomic number and density, and are commonly used to render pseudo-color maps for security screening~\cite{benedykciuk2021material,khan2020review,macdonald2001design,rebuffel2007dual}. 

In contrast, dual-view or multi-view systems improve geometric coverage by imaging baggage from multiple projection angles, but do not by themselves provide material decomposition unless combined with spectral sensing. 
CT-based EDS further reconstructs volumetric structure and can provide stronger three-dimensional analysis, at the cost of more specialized hardware and deployment constraints, while backscatter imaging offers complementary surface-sensitive contrast but is not a replacement for transmission-based high-energy and low-energy measurements.
Thus, dual-energy channels provide valuable physics-grounded material cues when available, but their accessibility is scanner-dependent rather than universal. 
In our PXISL module, using dual-energy maps allows Beer--Lambert composition to be performed independently in the high- and low-energy absorbance domains, preserving material-dependent attenuation during synthesis. When only single-energy X-ray scans are available, Eqs.~\ref{eq:absorbance_single_density_map} and ~\ref{eq:beer_lambert_add} naturally reduce to a single-channel density-map-based synthesis by applying the same absorbance-domain addition to the available attenuation map.

\begin{table}[t!]
    \centering
    \caption{
        Computational cost and curriculum overhead on eight NVIDIA
        H100 GPUs (80\,GB).
        The single-stage control uses the merged 18,000 synthetic scenes
        with an approximately matched number of optimization updates.
        Peak memory is reported per GPU.
        Clean wall time excludes measured startup overhead,
        including data-loader worker initialization.
        These measurements cover model training only.
    }
    \label{tab:computational_cost}
    \begin{adjustbox}{width=1.0\linewidth}
        \begin{tabular}{lcccccccc}
            \toprule
            Stage &
            \shortstack{Objects\\per image} &
            \shortstack{Images\\per epoch} &
            Epochs &
            Updates &
            \shortstack{Peak memory\\per GPU (GB)} &
            \shortstack{Startup\\time (h)} &
            \shortstack{Total wall\\time (h)} &
            \shortstack{Clean wall\\time (h)} \\
            \midrule
            Easy
                & 2--4 & 4,000 & 80 & 20,000
                & 22 & 0.62 & 2.70 & 2.08 \\
            Hard
                & 4--7 & 6,000 & 64 & 24,000
                & 22 & 0.49 & 3.47 & 2.98 \\
            Extreme
                & 6--9 & 8,000 & 48 & 24,000
                & 22 & 0.38 & 3.47 & 3.09 \\
            \midrule
            Full curriculum
                & 2--9 & 18,000 & $192^{\dagger}$ & 68,000
                & 22 & 1.49 & 9.64 & 8.15 \\
            Single-stage
                & 2--9 & 18,000 & 60 & 67,500
                & 22 & 0.48 & 8.94 & 8.46 \\
            \bottomrule
        \end{tabular}%
    \end{adjustbox}
    \par\smallskip
    \begin{minipage}{\linewidth}
        \footnotesize
        $\dagger$ The curriculum epoch count is the sum of the
        stage-specific epochs over different training sets,
        rather than 192 epochs over all 18,000 scenes.
    \end{minipage}
\end{table}

\noindent\textbf{Computational Cost and Curriculum Overhead.}
As shwon in~\cref{tab:computational_cost}, we compare the Easy → Hard → Extreme curriculum with single-stage training on the merged synthetic dataset. 
Both settings use the same model architecture, loss functions, global batch size, and hardware, with approximately matched training budgets of 68k and 67.5k updates, respectively. 
The complete curriculum requires 9.64 hours, corresponding to 77.12 H100 GPU-hours, compared with 8.94 hours for single-stage training. 
This represents an additional 0.70 hours, or approximately 7.8\% of wall-clock time. 
The difference is primarily attributable to startup overhead accumulated over more stage-specific epochs: 1.49 hours for the curriculum versus 0.48 hours for the single-stage control. 
After subtracting this overhead, the measured training times are comparable, at 8.15 and 8.46 hours. 
Peak GPU memory remains 22 GB in both settings. 
Thus, the curriculum introduces a modest scheduling overhead without changing the model architecture or increasing peak memory requirements.
These measurements cover model training; object-bank construction and offline scene synthesis are outside the scope of this comparison.

\begin{table}[t!]
  \centering
  \caption{Ablation study of pseudo-labeling granularity. We evaluate the performance of holistic instance masks, over-segmented local parts, and the proposed unified LAO-X strategy across five X-ray benchmarks.}
  \begin{adjustbox}{width=1.0\linewidth}
    \begin{tabular}{l|cc >{\columncolor{cyan!50!green!20}}c |cc >{\columncolor{cyan!50!green!20}}c |cc >{\columncolor{cyan!50!green!20}}c |cc >{\columncolor{cyan!50!green!20}}c |cc >{\columncolor{cyan!50!green!20}}c}
      \toprule
      \multirow{2}{*}{Labeling Strategy} & \multicolumn{3}{c|}{\textbf{CLCXray}} & \multicolumn{3}{c|}{\textbf{DvXray}} & \multicolumn{3}{c|}{\textbf{PIXray}}& \multicolumn{3}{c|}{\textbf{HiXray}} &  \multicolumn{3}{c}{\textbf{PIDray}} \\
      & AP$_{50}$ & AP$_{75}$ & AP$^{95}_{50}$ & AP$_{50}$ & AP$_{75}$ & AP$^{95}_{50}$ & AP$_{50}$ & AP$_{75}$ & AP$^{95}_{50}$ & AP$_{50}$ & AP$_{75}$ & AP$^{95}_{50}$ & AP$_{50}$ & AP$_{75}$ & AP$^{95}_{50}$ \\
      \midrule
      \multicolumn{16}{l}{Standard COCO evaluation} \\
      \midrule
      Holistic Object       
      & 17.46 & 13.65 & 12.24 
      & 3.97 & 2.91 & 3.16 
      & \underline{15.04} & \underline{11.48} & \underline{11.04} 
      & 16.73 & 8.74 & 8.83 
      & \textbf{6.49} & \textbf{2.98} & \underline{2.72} \\
      
      Partial Object        
      & 18.72 & 15.57 & 13.23 
      & \textbf{5.05} & \textbf{4.59} & \textbf{4.02} 
      & 8.81 & 6.41 & 6.52 
      & \underline{26.03} & \underline{13.57} & \underline{13.66} 
      & \underline{5.89} & 2.40 & \textbf{2.80} \\

      % RGB                   
      % & 25.78 & 20.69 & 18.05 
      % &  5.31 &  4.12 &  3.87 
      % & 15.63 & 11.63 & 11.32 
      % & 28.27 & 14.50 & 14.93 
      % &  5.94 &  3.00 &  3.17 \\
      
      \textbf{\method}  
      & \textbf{23.90} & \textbf{18.40} & \textbf{16.16} 
      & \underline{4.81} & \underline{4.32} & \underline{3.77} 
      & \textbf{16.03} & \textbf{11.93} & \textbf{11.38} 
      & \textbf{26.30} & \textbf{13.67} & \textbf{13.80} 
      & 5.42 & \underline{2.48} & 2.68  \\
      \bottomrule
    \end{tabular}
  \end{adjustbox}
\label{ablation_holistic_part_ap}
\end{table}

\section{Partial-Annotation Evaluation Protocol and Sensitivity Analysis}
\label{app:protocol}
Protocol definition. Several X-ray benchmarks annotate only selected target categories, leaving other objects unlabeled. 
Standard full-image evaluation can therefore count valid predictions of unlabeled objects as false positives. 
To reduce this effect, our region-filtered protocol retains a predicted box only if its center lies within at least one expanded ground-truth box. 
Each ground-truth box is expanded about its center by multiplying its width and height by a factor $\alpha$, with $\alpha$ = 1.5 in the main experiments. 
COCO matching and metric aggregation are then applied without modification. 
This procedure evaluates localization around annotated targets and should be distinguished from unrestricted full-image evaluation, since predictions outside the retained regions do not contribute to the reported metrics.
Sensitivity to region filtering. We evaluate SAM2, SAM3, UnSAMv2, and LAO-X under standard full-image evaluation and three expansion factors, $\alpha\in$ {1.25, 1.5, 1.75}. 
LAO-X achieves the highest AP$_{50}$ and AP$^{95}_{50}$ in all 16 dataset–metric–protocol combinations. 
Relative to the default protocol, the complete ordering of the four methods is preserved in 15 of these 16 combinations. Nevertheless, the protocol affects both absolute scores and improvement margins. 
On CLCXray, LAO-X exceeds the strongest baseline by 4.91 AP$^{95}_{50}$ points under full-image evaluation, compared with 6.23 points under the default 1.5× filter.
Across the three expansion factors, its AP$^{95}_{50}$ advantage ranges from 5.52 to 6.23 points. 
These results show that LAO-X’s leading performance among the evaluated segmenters is not specific to the default filter, while the magnitude of its advantage remains protocol-dependent.
Unmatched predictions and false discoveries. 
On CLCXray, at a score threshold of 0.9 and matching IoU of 0.5, the default filter removes 70.8\%–76.2\% of unmatched predictions across the four methods. 
Because annotations are incomplete, these removed predictions may include both genuine errors and correctly localized but unlabeled objects. 
Their removal should therefore not be interpreted as a measured reduction in the true false-positive rate. 
We additionally compare LAO-X and SAM2 at a matched false discovery rate on DET-COMPASS, defining FDR = FP/(TP + FP). At an FDR of 70.42\%, LAO-X recalls 85.11\% of annotated objects, compared with 54.37\% for SAM2. 
This operating-point comparison provides complementary evidence that LAO-X recovers more annotated objects at the same proportion of false discoveries. 
Partial annotations nevertheless prevent a complete assessment of full-image errors on datasets such as CLCXray.

\begin{table}[t]
    \centering
    \caption{
        Sensitivity to the partial-annotation evaluation protocol
        on DET-COMPASS and CLCXray.
        Each entry reports box
        AP$_{50}$ / AP$^{95}_{50}$.
        Full-image evaluation applies standard COCO evaluation
        without region filtering.
        The other columns retain predictions whose box centers
        lie within at least one ground-truth box expanded by the
        indicated factor.
        The $1.5\times$ setting is used in the main comparison.
        The best result in each column within each dataset is
        shown in bold.
    }
    \label{tab:protocol_sensitivity}
    \begin{adjustbox}{width=0.6\linewidth}
        \begin{tabular}{lcccc}
            \toprule
            Method &
            Full image &
            $1.25\times$ &
            $1.5\times$ (main) &
            $1.75\times$ \\
            \midrule
            \multicolumn{5}{l}{\textit{DET-COMPASS}} \\
            \midrule
            SAM2
                & 26.23 / 22.43
                & 37.86 / 31.01
                & 36.10 / 29.74
                & 34.58 / 28.62 \\
            SAM3
                & 7.87 / 6.90
                & 13.37 / 11.18
                & 11.46 / 9.69
                & 10.39 / 8.86 \\
            UnSAMv2
                & 18.47 / 16.25
                & 37.49 / 30.62
                & 33.25 / 27.44
                & 30.11 / 25.07 \\
            LAO-X
                & \textbf{55.39 / 46.91}
                & \textbf{64.86 / 54.46}
                & \textbf{63.39 / 53.33}
                & \textbf{62.62 / 52.75} \\
            \midrule
            \multicolumn{5}{l}{\textit{CLCXray}} \\
            \midrule
            SAM2
                & 8.33 / 6.05
                & 20.68 / 14.90
                & 17.75 / 12.80
                & 15.88 / 11.47 \\
            SAM3
                & 12.42 / 8.47
                & 33.07 / 22.20
                & 28.14 / 18.93
                & 24.70 / 16.65 \\
            UnSAMv2
                & 17.11 / 11.25
                & 34.36 / 21.15
                & 30.71 / 19.06
                & 28.29 / 17.70 \\
            LAO-X
                & \textbf{23.90 / 16.16} %\textbf{18.39 / 12.62}
                & \textbf{42.06 / 28.07}
                & \textbf{37.61 / 25.29}
                & \textbf{34.33 / 23.22} \\
            \bottomrule
        \end{tabular}%
    \end{adjustbox}
\end{table}

\section{Segmentation evaluation on real X-ray inspection dataset}
\label{app:seg}

\begin{table}[h]
  \centering
  \caption{Segmentation evaluation results on PIXray and PIDray (using standard COCO metric).}
  \begin{adjustbox}{width=0.8\linewidth}
    \begin{tabular}{l|cc ccc|cc ccc}
      \toprule
      \multirow{2}{*}{\textbf{Method}} & \multicolumn{5}{c|}{\textbf{PIXray}} &  \multicolumn{5}{c}{\textbf{PIDray}} \\
      & AR$_1$ & AR$_{10}$  & AP$_{50}$ & AP$_{75}$ & AP$^{95}_{50}$ & AR$_1$ & AR$_{10}$ & AP$_{50}$ & AP$_{75}$ & AP$^{95}_{50}$ \\
      \midrule
      SAM2              &  0.64 & 20.42 &  3.48 &  3.22 &  2.85 &  3.18 &  8.19 &  0.89 &  0.53 &  0.51\\
      SAM3              &  0.60 & 29.74 &  6.54 &  5.41 &  4.88 &  2.24 & \underline{21.62} &  2.57 &  1.04 &  1.19\\
      UnSAMv2           & \textbf{10.70} & \underline{29.20} & \underline{13.60} & \underline{11.02} & \underline{10.41} & \textbf{5.81} & 15.61 &  \textbf{3.80} &  \underline{1.59} &  \underline{1.83} \\
      \textbf{\method}  & 6.85 & \textbf{41.94} & \textbf{15.29} & \textbf{11.80} & \textbf{10.93} & \underline{4.35} & \textbf{24.14} &  \underline{3.79} &  \textbf{1.78} &  \textbf{1.88}\\
      \bottomrule
    \end{tabular}
  \end{adjustbox}
\label{segmentation_results_apar}
\end{table}
We select PIXray and PIDray to evaluate segmentation since they provide pixel-level masks for hazardous objects.
Tab.~\ref{segmentation_results_apar} reports COCO-style mask metrics.
Across both datasets, \method delivers the best overall segmentation quality: it achieves the highest mIoU / F1 on PIXray (64.06\% / 71.86\%) and PIDray (50.37\% / 61.17\%), outperforming UnSAMv2 by large margins. COCO metrics further show that \method improves object coverage under clutter, reaching the best AR$_{10}$ on both PIXray (41.94\% vs.\ 29.20\%) and PIDray (24.14\% vs.\ 15.61\%). On PIXray, \method also yields the highest precision (AP$_{50}$: 15.29\%, AP$^{95}_{50}$: 10.93\%). On PIDray, \method remains comparable on AP$_{50}$ (3.79\% vs.\ 3.80\%) while improving stricter metrics (AP$_{75}$: 1.78\% vs.\ 1.59\%; AP$^{95}_{50}$: 1.88\% vs.\ 1.83\%), indicating more accurate masks when localized. Overall, the consistent gains in both region-based and COCO-style metrics confirm that \method produces cleaner masks with higher coverage on real-world X-ray scans.

\section{Pseudo label of holistic/partial objects and their combination}
\label{app:pseudo}
To evaluate the impact of multi-granularity supervision, we compare training with holistic masks derived from saliency-guided mining (Holistic), over-segmented local components (Part), and our combined strategy (\method). 
As shown in Tab.~\ref{ablation_holistic_part_ap}, \method significantly outperforms holistic or part pseudo-labels across most benchmarks. 
For instance, on CLCXray, \method achieves 23.90\% AP$_{50}$, markedly higher than Holistic (17.46\%) or Part (18.72\%) alone. 
This improvement is even more evident in HiXray, where AP$_{50}$ increases from 16.73\% to 26.30\% (+9.57).
These results demonstrate that holistic and part-level annotations provide complementary information: while part labels handle internal complexity, for example in DvXray (5.05\% AP$_{50}$), our integrated approach ensures the most robust and precise localization across various distribution of X-ray targets. 
By effectively merging these multi-granularity cues, \method learns the complex, multi-overlapped structures inherent in security scans.

\section{Synthetic Scene Complexity and Data Scale}
\label{app:comp}
\noindent\textbf{Occlusion-Controlled Curriculum.}
Our occlusion-controlled curriculum learning strategy (\cref{sec:curriculum}) consists of Easy~/~Hard~/~Extreme stages, increasing both the number of pasted items $N$ of $\{[2,4]~/~[4,7]~/~[6,9]\}$ and the overlap level $\gamma$ of $\{[0.0,0.05]~/~[0.0,0.3]~/~[0.0,0.5]\}$, where the learning rate and training epochs is set for \{$1.6 \times 10^{-4}~/~1.2 \times 10^{-4}~/~1.0 \times 10^{-4}$\}, and $\{80~/~64~/~48\}$.
We train sequentially by stages and reset the optimizer at the start of each stage.

\noindent\textbf{Sensitivity to Object Count and Overlap.}
Moreover, we evaluate framework sensitivity to synthesis complexity by varying the object count ($N$) and overlap ratio ($\gamma$) during the extreme stage. 
As illustrated in Tab.~\ref{ablation_curriculum_synthetic_ap}, performance remains remarkably stable across these configurations, with $AP$ metrics fluctuating only marginally as $N$ increases up to $15$ or $\gamma$ expands to $0.9$.
Notably, increasing $N$ and $\gamma$ naturally provides a higher volume of both holistic and part annotations for training. 
This indicates that leveraging a denser set of supervision signals has the potential to further improve model performance.
However, comparing these results with the ablation of curriculum tiers in~\cref{ablation} (Main paper) of curriculum learning reveals that the staged design of the curriculum is a much more significant factor than specific stage hyper-parameters.
This strengthen our conclusion that the sequential transition from simple to heavily cluttered scenes is the primary driver for optimization stability and generalization in the projective X-ray domain.

\begin{table}[t!]
  \centering
  \caption{Sensitivity analysis of synthesis parameters. The table reports performance variations across different object counts ($N$) and overlap ratios ($\gamma$) within the extreme curriculum stage.}
  \begin{adjustbox}{width=1.0\linewidth}
    \begin{tabular}{cc|cc >{\columncolor{cyan!50!green!20}}c cc|cc >{\columncolor{cyan!50!green!20}}c cc|cc >{\columncolor{cyan!50!green!20}}c cc|cc >{\columncolor{cyan!50!green!20}}c cc|cc >{\columncolor{cyan!50!green!20}}c cc|cc >{\columncolor{cyan!50!green!20}}c cc}
      \toprule
      \multicolumn{2}{c|}{Synthetic parameter} & \multicolumn{5}{c|}{\textbf{CLCXray}} & \multicolumn{5}{c|}{\textbf{DvXray}} &  \multicolumn{5}{c|}{\textbf{PIXray}} & \multicolumn{5}{c|}{\textbf{HiXray}} & \multicolumn{5}{c}{\textbf{PIDray}} & \multicolumn{5}{c}{\textbf{Average}} \\
      Objects ($N$) & Overlap ($\gamma$) & AP$_{50}$ & AP$_{75}$ & AP$^{95}_{50}$ & AR$_1$ & AR$_{10}$ 
      & AP$_{50}$ & AP$_{75}$ & AP$^{95}_{50}$ & AR$_1$ & AR$_{10}$ 
      & AP$_{50}$ & AP$_{75}$ & AP$^{95}_{50}$ & AR$_1$ & AR$_{10}$ 
      & AP$_{50}$ & AP$_{75}$ & AP$^{95}_{50}$ & AR$_1$ & AR$_{10}$ 
      & AP$_{50}$ & AP$_{75}$ & AP$^{95}_{50}$ & AR$_1$ & AR$_{10}$
      & AP$_{50}$ & AP$_{75}$ & AP$^{95}_{50}$ & AR$_1$ & AR$_{10}$ \\
      \midrule
      (\method) 6-9 & [0.0,0.5]
      & \underline{37.61} & \underline{27.44} & \underline{25.29} & \underline{28.63} & \underline{50.20}
      & 23.56 & 18.33 & 16.87 & 30.54 & 46.44
      & \textbf{30.26} & \textbf{22.40} & \textbf{21.37} & 13.52 & \textbf{51.86}
      & 43.35 & 21.75 & 22.47 & 16.70 & 34.97
      & \textbf{11.61} & 5.16 & 5.65 & 12.54 & 37.06 
      & 29.28 & 19.01 & 18.33 & \underline{20.38} & 44.11 \\	
      9-12  & [0.0,0.5]
      & 37.38 & 27.42 & 25.17 & 27.83 & 49.78 
      & \underline{23.89} & \underline{18.79} & \underline{17.26} & \underline{31.11} & \underline{46.79} 
      & 29.79 & 21.93 & 20.97 & 13.09 & 51.60 
      & \textbf{44.57} & \underline{22.30} & \textbf{23.10} & \textbf{17.00} & \textbf{35.64} 
      & 11.29 &  5.20 &  5.62 & 12.52 & 37.10	
      & \underline{29.38} & \underline{19.13} & \underline{18.42} & 20.31 & \underline{44.18} \\ 
      12-15 & [0.0,0.5]
      & \textbf{39.01} & \textbf{29.30} & \textbf{26.46} & \textbf{28.77} & \textbf{51.77} 
      & \textbf{24.25} & \textbf{18.94} & \textbf{17.52} & \textbf{32.00} & \textbf{47.99}
      & \underline{30.13} & \underline{22.16} & \underline{21.20} & \textbf{13.81} & \underline{51.81} 
      & 43.86 & 21.49 & 22.63 & 16.78 & 35.11 
      & 11.33 &  \underline{5.37} &  \underline{5.67} & \textbf{12.89} & \underline{37.46}
      & \textbf{29.71} & \textbf{19.45} & \textbf{18.70} & \textbf{20.85} & \textbf{44.83}	\\
      6-9   & [0.0,0.7]
      & 36.81 & 27.06 & 24.66 & 27.66 & 49.33 
      & 23.16 & 18.01 & 16.56 & 29.95 & 45.28 
      & 29.57 & 21.96 & 20.90 & 13.43 & 50.78 
      & \underline{44.40} & \textbf{22.31} & \underline{23.06} & \underline{16.93} & \underline{35.60} 
      & 10.76 &  4.89 &  5.30 & 12.17 & 36.37
      & 28.94 & 18.85 & 18.10 & 20.03 & 43.47	\\
      6-9   & [0.0,0.9]
      & 38.59 & 27.46 & 24.74 & 28.62 & 41.36 
      & 23.19 & 17.97 & 16.59 & 30.14 & 45.05 
      & 29.22 & 21.36 & 20.42 & 12.30 & 50.93 
      & 44.10 & 21.91 & 22.79 & 16.87 & 35.27 
      & \underline{11.39} &  \textbf{5.45} &  \textbf{5.78} & \underline{12.76} & \textbf{38.05}
      & 29.30 & 18.83 & 18.07 & 20.14 & 42.13\\
      \bottomrule
    \end{tabular}
  \end{adjustbox}
\label{ablation_curriculum_synthetic_ap}
\end{table}

\begin{table}[t]
    \centering
    \caption{
        Effect of synthetic data scale on CLCXray.
        We compare 4,500 and 18,000 synthetic scenes while preserving
        the Easy/Hard/Extreme proportions, stage-specific epoch counts,
        and learning-rate schedule.
        Results use the default $1.5\times$ region-filtered
        evaluation protocol.
        Since the number of epochs is fixed, the reduced-data
        configuration also uses fewer optimization updates.
        The best results are shown in bold.
    }
    \label{tab:synthetic_data_scale}
    \begin{adjustbox}{width=0.8\linewidth}
        \begin{tabular}{lcrrrrr}
            \toprule
            Training data &
            \shortstack{Easy / Hard / Extreme\\scenes} &
            AP$_{50}$ &
            AP${75}$ &
            AP$^{95}_{50}$ &
            AR$_{1}$ &
            AR$_{10}$ \\
            \midrule
            4,500 scenes (25\%)
                & 1,000 / 1,500 / 2,000
                & 34.19
                & 24.48
                & 22.46
                & 26.83
                & 40.27 \\
            18,000 scenes (100\%)
                & 4,000 / 6,000 / 8,000
                & \textbf{37.61}
                & \textbf{27.44}
                & \textbf{25.29}
                & \textbf{28.63}
                & \textbf{50.20} \\
            \bottomrule
        \end{tabular}%
    \end{adjustbox}
\end{table}

\noindent\textbf{Effect of Synthetic Data Scale.} 
We examine the effect of reducing the synthetic training set from 18,000 to 4,500 scenes. 
The reduced setting uses 1,000, 1,500, and 2,000 scenes in the Easy, Hard, and Extreme stages, respectively, preserving their relative proportions. 
We retain the same synthesis parameters, model configuration, stage-specific epoch counts, and learning-rate schedule. 
On CLCXray, the reduced setting achieves 22.46 AP$^{95}_{50}$, compared with 25.29 for the full dataset, a decrease of 2.83 points. 
AP$_{50}$ decreases from 37.61 to 34.19, while AR$_{10}$ decreases from 50.20 to 40.27, indicating a larger reduction in object coverage when multiple predictions are allowed. 
These results show that the full training configuration benefits from the larger synthetic dataset, while the reduced configuration retains substantial localization performance. 
Since the number of epochs is fixed, reducing the dataset also reduces the number of optimization updates. 
This experiment therefore measures the effect of data scale under a fixed-epoch schedule, rather than isolating data diversity under a fixed computational budget.

\begin{figure}[t]
    \centering
    \includegraphics[width=\linewidth]{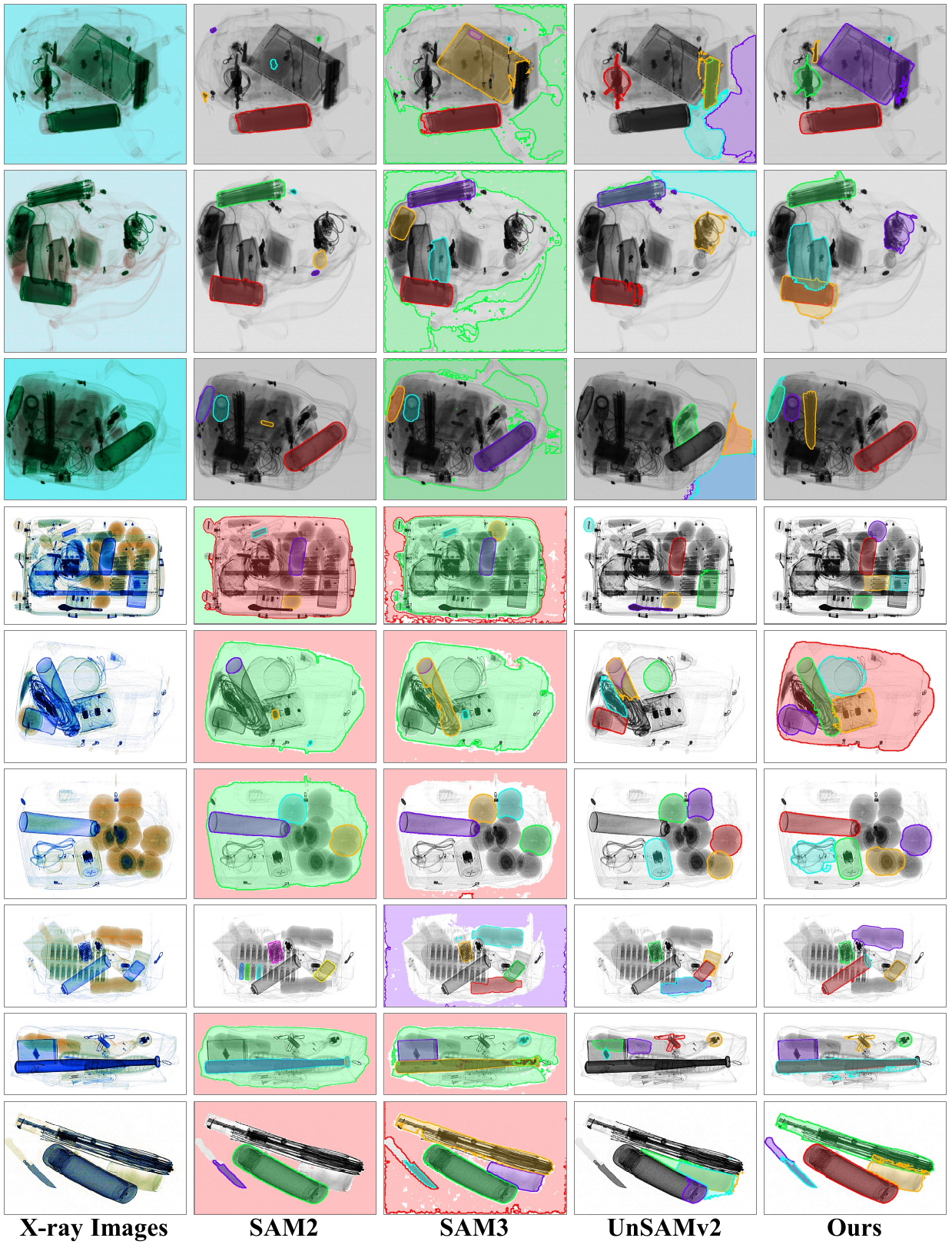}
    \caption{\textbf{More on comparison of masks with top-5 confidence across real-world X-ray datasets.}
    To highlight the predicted mask, we convert the original image to a grayscale image in our visualizations.}
    \label{fig:top5_mask_app_1}
\end{figure}

\begin{figure}[t]
    \centering
    \includegraphics[width=\linewidth]{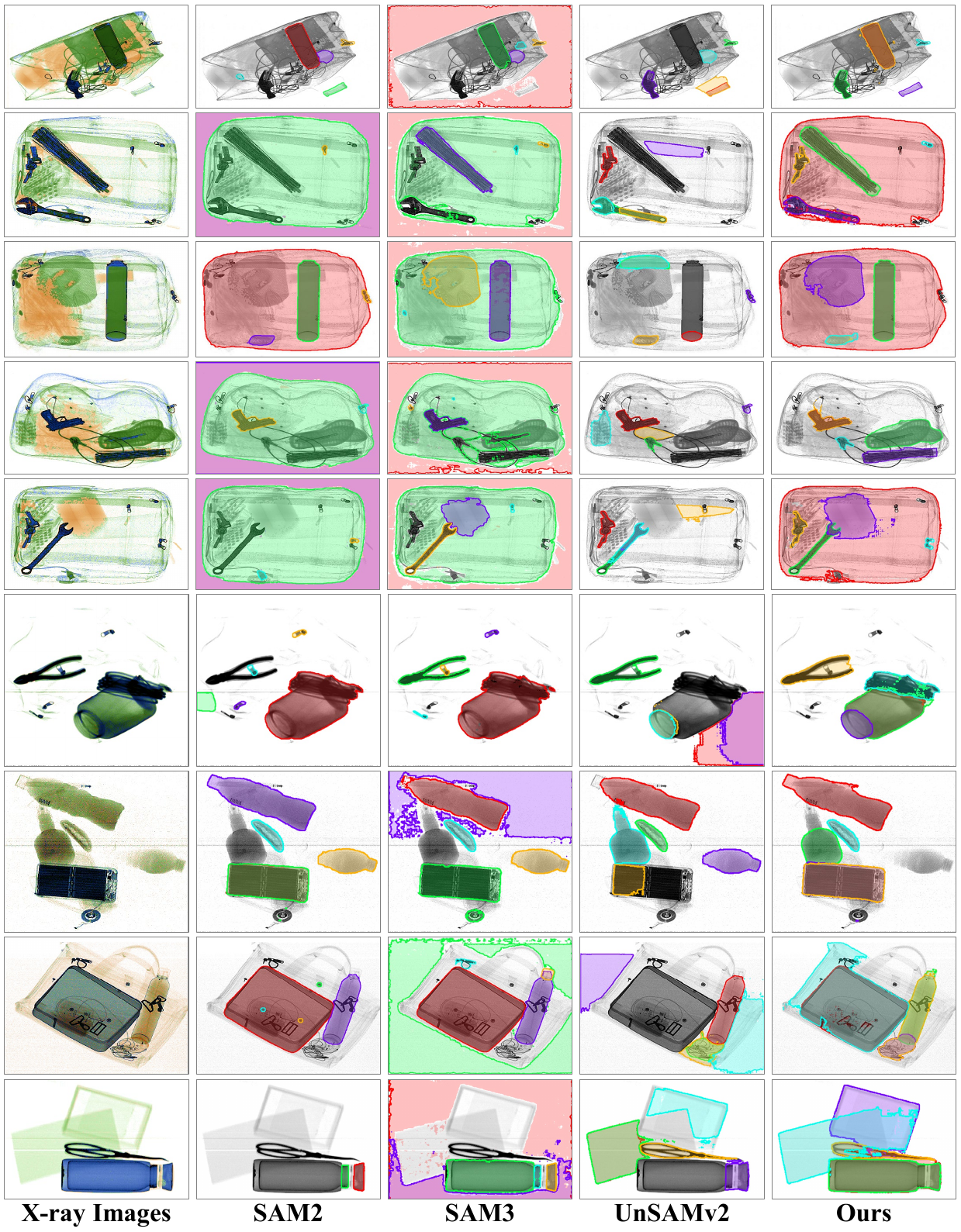}
    \caption{\textbf{More on comparison of masks with top-5 confidence across real-world X-ray datasets.}
    To highlight the predicted mask, we convert the original image to a grayscale image in our visualizations.}
    \label{fig:top5_mask_app_2}
\end{figure}

\section{Qualitative Comparison of Mask Localization Performance.}
\label{app:mask_crop}

\cref{fig:top5_mask_app_1,fig:top5_mask_app_2} provide additional top-5 confident mask predictions across real-world X-ray datasets, where the original scans are converted to grayscale to make the predicted masks easier to inspect. 
SAM2 often produces sparse or incomplete masks, especially for low-contrast and overlapped objects, while SAM3 tends to generate larger but noisy regions that frequently spill into background or merge nearby structures. 
UnSAMv2 improves the objectness of predicted regions, but its mask extents remain unstable under heavy superposition and complex baggage layouts. 
In contrast, \method produces more object-aligned top-confidence masks, with cleaner boundaries, better instance separation, and stronger preservation of fine structures. 
These visualizations also show that \method can recover plausible objects beyond the manually annotated prohibited items, supporting its ability to learn a more general X-ray localization prior rather than only fitting labeled target categories.

\section{Quantitative and Qualitative Analysis of Interactive Multi-Granularity Segmentation.}
\label{app:interactive}
\noindent\textbf{Quantitatively analysis:}
Fig.~\ref{fig:granularity_sweep} analyzes how the queried granularity \(g\in[0.1,1.0]\) affects localization accuracy of \method.
On DET-COMPASS, AP$_{50}$ increases almost monotonically with \(g\), indicating that larger granularity encourages holistic masks that better align with fully annotated boxes.
On real-world prohibited-item benchmarks, the optimal \(g\) is dataset-dependent: PIDray favors coarse predictions (best around  $g\approx0.8$), while on PIXray, performance varies non-monotonically with granularity, with a local peak around \(g=0.3\) and the highest AP$_{50}$ at \(g=1.0\)
Though the inherent distribution of object scale shifts across different datasets, the proposed methods maintain strong capability at granularity 1.0.
Overall, the sweep confirms that granularity is a practical control knob: \method can adapt its output scale to different datasets and labeling regimes, and we report the optimal \(g\) for each benchmark in all quantitative comparisons.

\begin{figure}
    \centering
    \includegraphics[width=\linewidth]{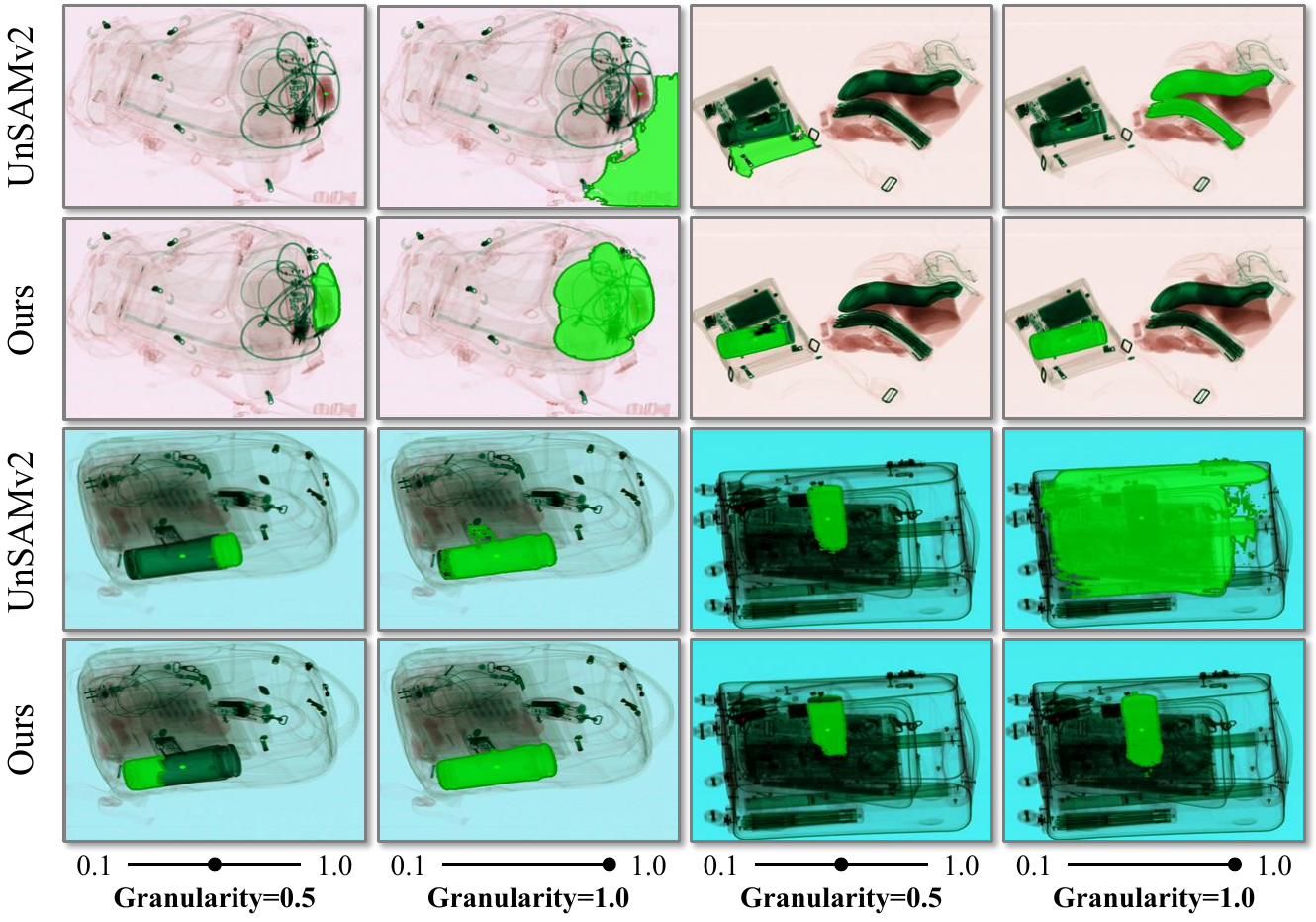}
    \label{fig:interactive_visualization_clcxray}
    \caption{\textbf{Granularity comparison with UnSAMv2 on CLCXray.} Green overlays are predicted masks at the queried granularity $g$ (shown under each column; same image prompt, only $g$ changes).
   UnSAMv2 is unstable and may collapse to over-segmenting large regions, while \method follows $g$ more consistently, transitioning from part-level (smaller $g$) to holistic masks (larger $g$) under heavy overlap.}
    \label{fig:interactive_vis_suppl_clcxray}
\end{figure}

\begin{figure}[t]
  \centering
   \includegraphics[width=1\linewidth]{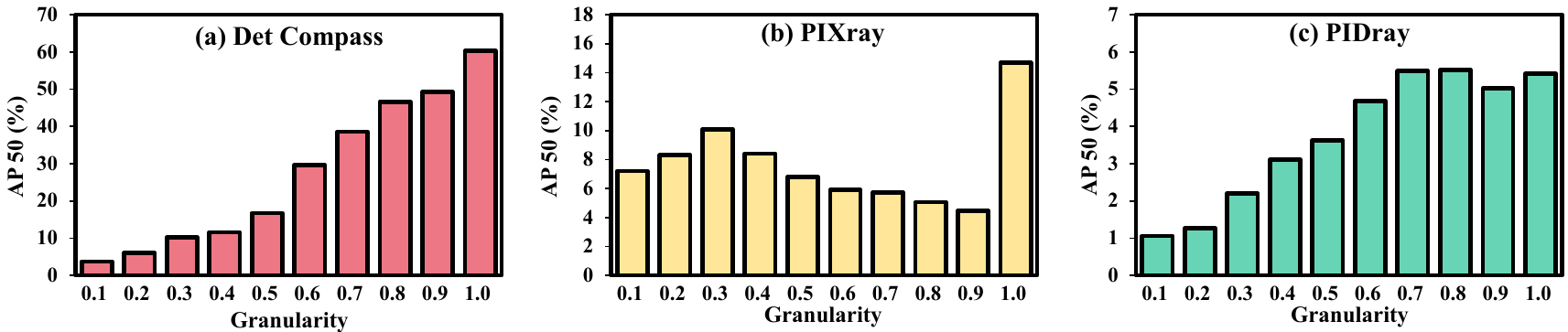}
\caption{\textbf{Granularity sweep of \method.} AP$_{50}$ as a function of the queried granularity $g\in[0.1,1.0]$ on DET-COMPASS and real-world X-ray benchmarks.}
\label{fig:granularity_sweep}
\end{figure}

\begin{figure}[htbp]
    \centering

    \begin{subfigure}{0.75\textwidth}
        \centering
        \includegraphics[width=\linewidth]{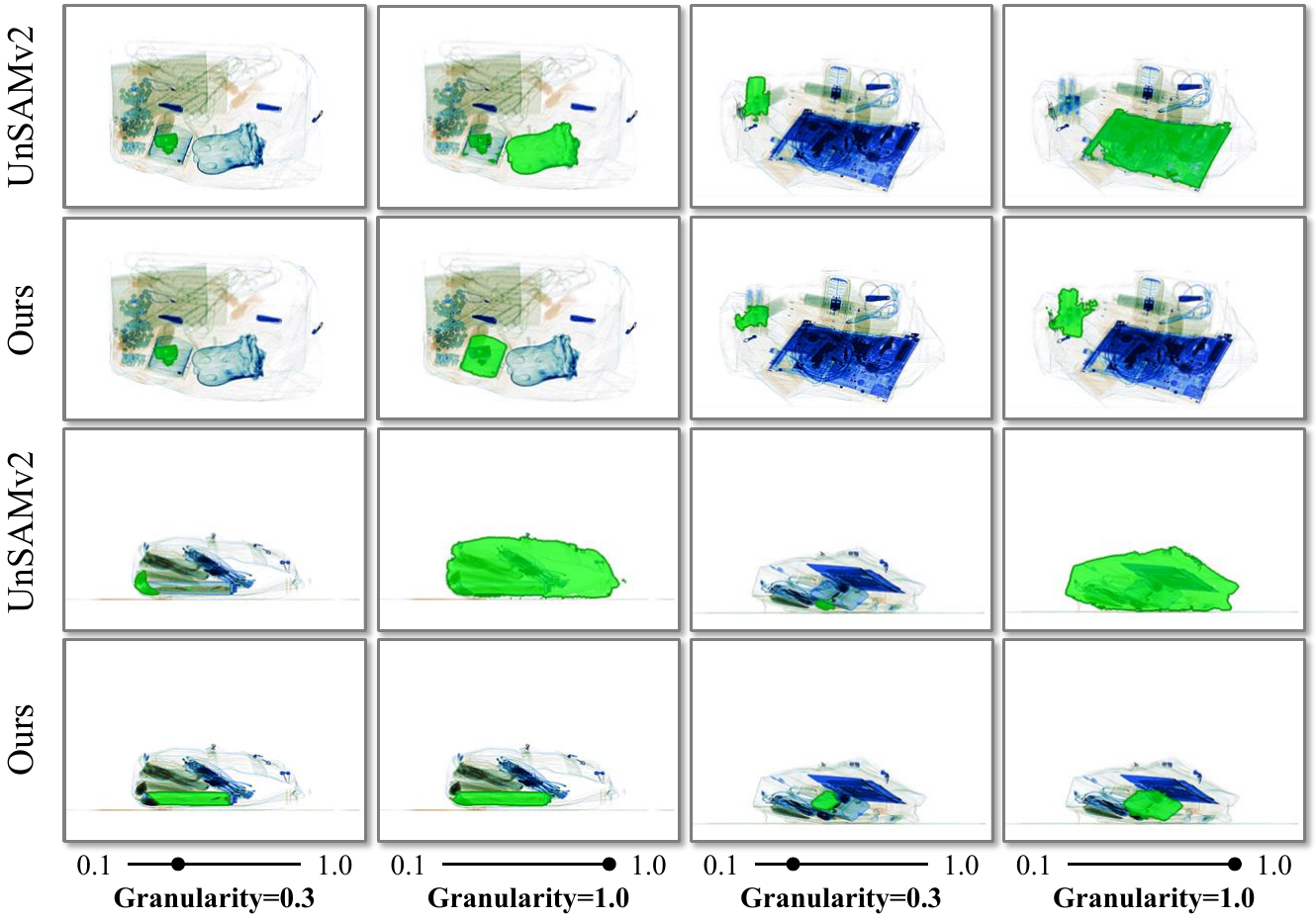}
        \label{fig:interactive_visualization_dvxray}
        \caption{DvXray}
    \end{subfigure}

    \begin{subfigure}{0.75\textwidth}
        \centering
        \includegraphics[width=\linewidth]{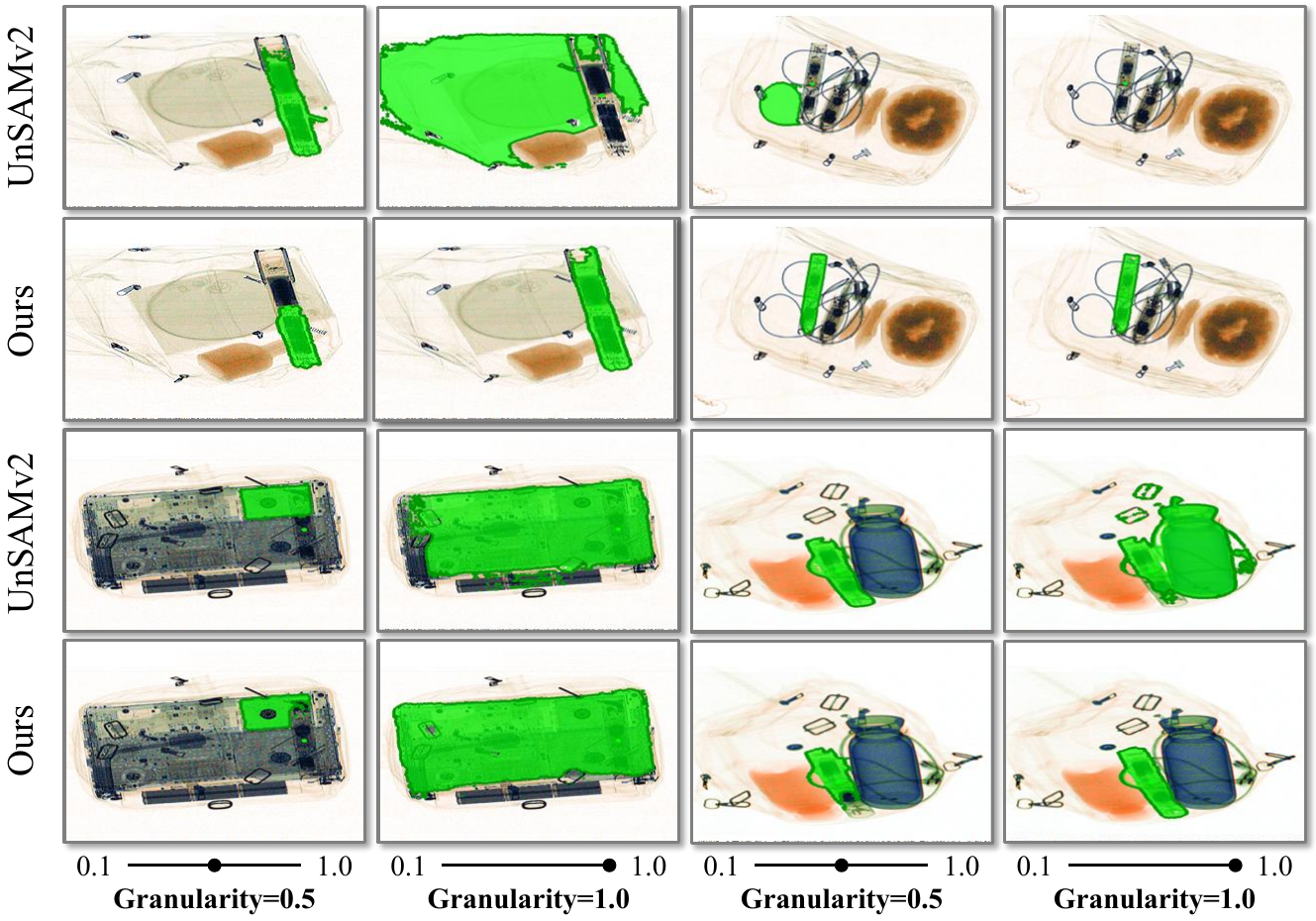}
        \label{fig:interactive_visualization_hixray}
        \caption{HiXray}
    \end{subfigure}
    \caption{More on \textbf{Granularity comparison with UnSAMv2 on DvXray and HiXray.}}
    \label{fig:interactive_vis_suppl_1}
\end{figure}

\begin{figure}[htbp]
    \centering

    \begin{subfigure}{0.75\textwidth}
        \centering
        \includegraphics[width=\linewidth]{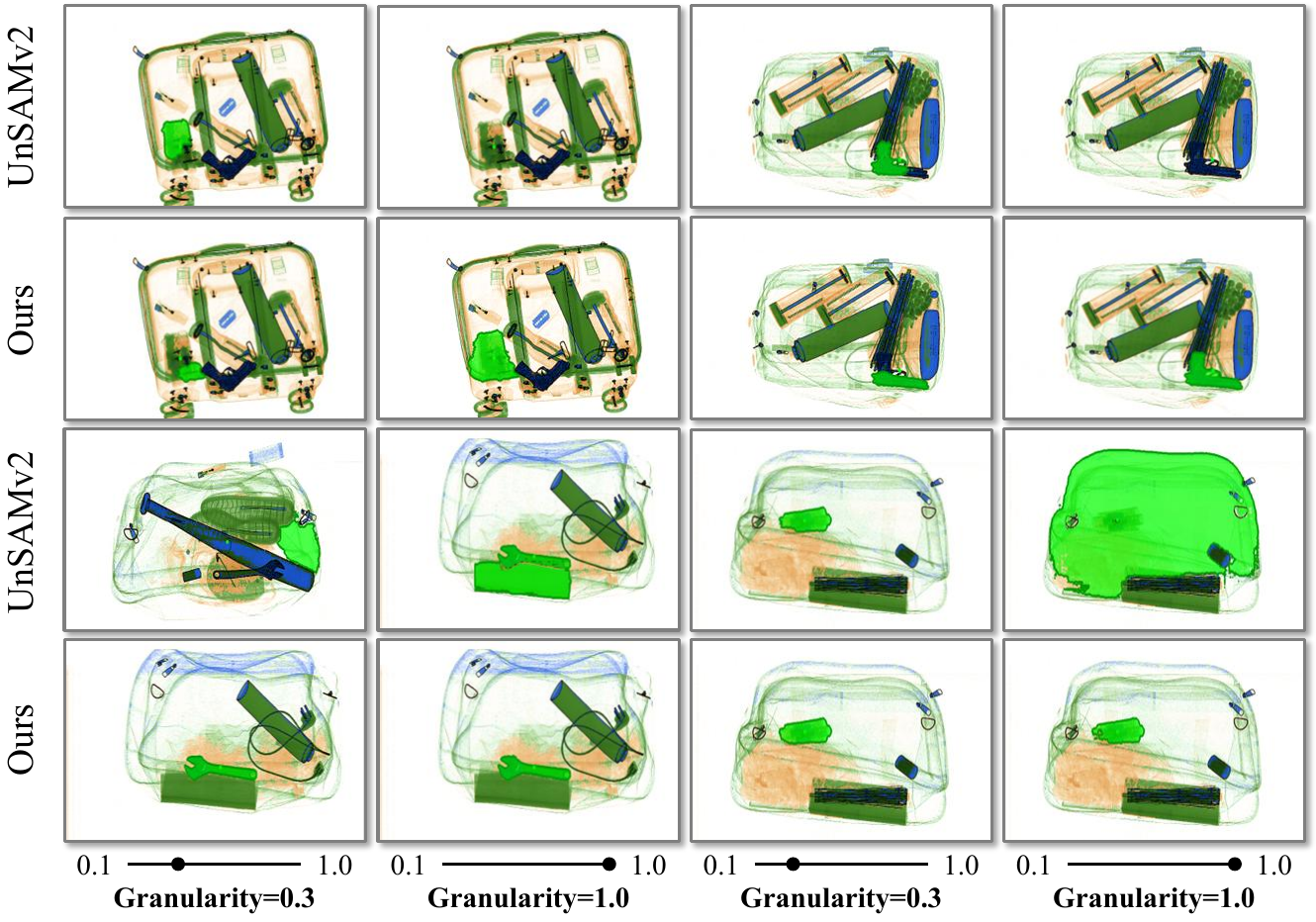}
        \label{fig:interactive_visualization_pixray}
        \caption{PIXray}
    \end{subfigure}
    
    \begin{subfigure}{0.75\textwidth}
        \centering
        \includegraphics[width=\linewidth]{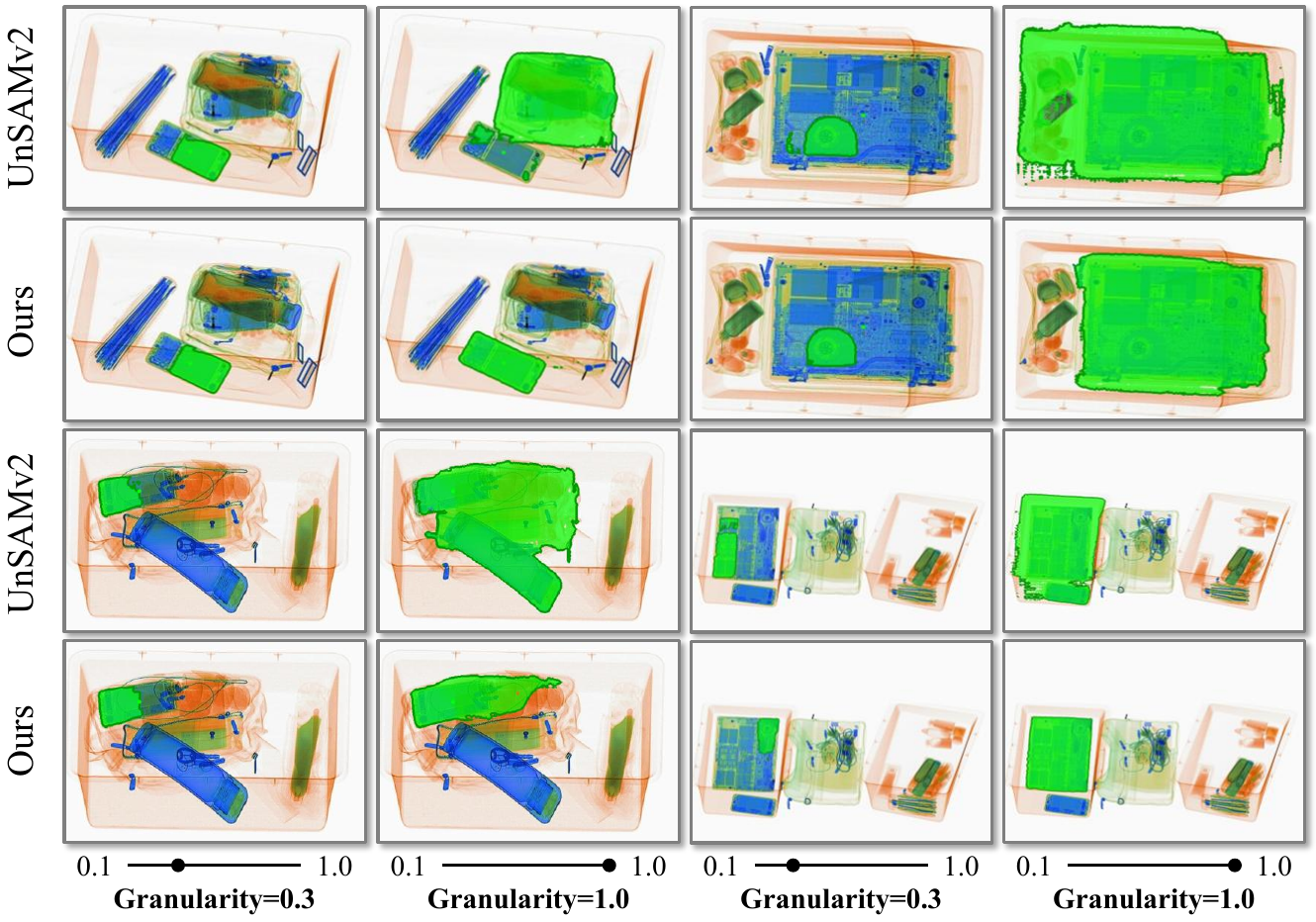}
        \label{fig:interactive_visualization_pidray}
        \caption{PIDray}
    \end{subfigure}
    \caption{More on \textbf{Granularity comparison with UnSAMv2 on PIXray and PIDray.}}
    \label{fig:interactive_vis_suppl_2}
\end{figure}

\noindent\textbf{Qualitative analysis:}
We evaluate the interactive segmentation capabilities of \method by querying images with the same point prompt across different granularity values $g$. 
As shown in~\cref{fig:interactive_vis_suppl_clcxray,fig:interactive_vis_suppl_1,fig:interactive_vis_suppl_2}, UnSAMv2 exhibits unstable behavior where changing $g$ often triggers incorrect mask shifts, such as expanding into the background or failing to define distinct parts under overlap. 
In contrast, \method responds to $g$ in a stable and monotonic manner: a smaller $g$ produces compact part-level regions, while a larger $g$ progressively expands toward the holistic object mask. 
This consistency, even in cluttered scans, confirms that our multi-granularity supervision enables the model to learn reliable, scale-conditioned localization priors for X-ray security imagery.

\section{Failure Cases and Limitations}
\label{app:failure}

\noindent\textbf{Low-attenuation objects.}
As shwon in~\cref{tab:low_attenuation}, very weak attenuation remains a challenging regime for LAO-X. 
On the official invisible subset of DET-COMPASS, LAO-X achieves 26.69 AP$_{50}$, substantially below its 63.39 AP$_{50}$ on the full evaluation set. 
Objects made primarily of thin plastic, fabric, or paper may provide attenuation cues close to the background, making their boundaries difficult to distinguish. 
In such cases, LAO-X may produce fragmented masks or miss the object entirely. 
These results qualify the scope of category-agnostic localization: removing a predefined semantic vocabulary does not eliminate the difficulty of recovering objects with weak image evidence.

\begin{table}[t!]
    \centering
    \caption{
        Localization of low-attenuation objects.
        We report LAO-X's box $\mathrm{AP}_{50}$ on the full
        DET-COMPASS evaluation set and its official invisible subset.
        The lower performance on the invisible subset highlights
        the difficulty of localizing objects with extremely weak
        attenuation cues.
    }
    \label{tab:low_attenuation}
    \begin{tabular}{lc}
        \toprule
        Evaluation set & $\mathrm{AP}_{50}$ \\
        \midrule
        DET-COMPASS, full set         & 63.39 \\
        DET-COMPASS, invisible subset & 26.69 \\
        \bottomrule
    \end{tabular}
\end{table}

\begin{table}[t!]
    \centering
    \caption{
        Localization under extreme overlap on the PIDray Hidden subset.
        LAO-X achieves the highest box AP and $\mathrm{AR}_{1}$
        among the evaluated methods, while its lower
        $\mathrm{AR}_{10}$ than SAM3 and UnSAMv2 indicates
        incomplete recovery of heavily superposed instances.
        The best results are shown in bold.
    }
    \label{tab:extreme_overlap}
    \begin{tabular}{lrrrrr}
        \toprule
        Method &
        $\mathrm{AP}_{50}$ &
        $\mathrm{AP}_{75}$ &
        $\mathrm{AP}_{50:95}$ &
        $\mathrm{AR}_{1}$ &
        $\mathrm{AR}_{10}$ \\
        \midrule
        SAM2
            & 0.03
            & 0.02
            & 0.02
            & 0.34
            & 0.36 \\
        SAM3
            & 4.18
            & 2.13
            & 2.27
            & 7.74
            & \textbf{31.35} \\
        UnSAMv2
            & 3.84
            & 1.18
            & 1.61
            & 8.05
            & 19.59 \\
        LAO-X
            & \textbf{5.87}
            & \textbf{2.60}
            & \textbf{2.98}
            & \textbf{11.92}
            & 14.60 \\
        \bottomrule
    \end{tabular}
\end{table}

\noindent\textbf{Extreme overlap.}
As shwon in~\cref{tab:extreme_overlap}, on the PIDray Hidden subset, where target objects are heavily superposed with surrounding contents, LAO-X achieves the highest AP$_{50}$, AP$_{75}$, AP$^{95}_{50}$, and AR$_{1}$ among the evaluated methods. 
However, its AR$_{10}$ is 14.60, below SAM3 at 31.35 and UnSAMv2 at 19.59. Thus, stronger AP and top-ranked object coverage do not imply that LAO-X recovers all instances when more predictions are permitted. 
Typical errors include merging neighboring objects into a single region, predicting only a component of an object, and missing instances whose attenuation patterns are dominated by overlapping contents. 
The low absolute AP across all methods further highlights the difficulty of this setting.

Scope of synthesis and deployment. 
PXISL recomposes real X-ray object measurements and backgrounds in the absorbance domain, preserving the additive attenuation structure used by our synthesis model. 
It does not explicitly simulate scattering, beam hardening, detector noise, calibration differences, or scanner-specific pseudo-color mappings. 
Consequently, performance across the evaluated datasets does not guarantee robustness to every scanner or operating condition. 
Together with the failure cases above and the limitations of partial annotations, this motivates using LAO-X as an assistive localization front end for downstream recognition and human inspection. 
Practical use requires validation and calibration on the target scanner, human oversight, and monitoring of missed objects and other failure modes.

% \clearpage

\section{Broader Impact}
\label{app:broaderimpact}
The primary motivation of this work is to improve open-world object localization in X-ray security screening.
By adapting a promptable localization model to X-ray scans without manual X-ray training annotations, \method will support beneficial applications such as assisting human operators, improving inspection efficiency, and helping identify unusual or previously unseen objects in cluttered baggage.
Because \method learns from unlabeled X-ray data and physics-guided synthetic samples, it may reduce the need for expensive expert annotation and make X-ray perception research more scalable.

At the same time, X-ray security screening is a safety-critical application.
Incorrect predictions may have practical consequences: false negatives may miss dangerous items, while false positives may increase unnecessary manual inspection or delay passengers.
The method may also be affected by scanner differences, dataset bias, unusual object configurations, or changes in operating conditions.
For these reasons, \method should be used as an assistive tool rather than as the sole basis for security decisions.
Responsible deployment requires human oversight, careful validation on the target scanner and environment, clear communication of system uncertainty, and continuous monitoring for failure cases after deployment.

%%%%%%%%%%%%%%%%%%%%%%%%%%%%%%%%%%%%%%%%%%%%%%%%%%%%%%%%%%%%
% \clearpage
% \newpage
% \input{checklist.tex}

\end{document}